\documentclass[twoside,11pt]{article}
\input{latexdefs}

\jmlrheading{1}{2016}{1-48}{4/00}{10/00}{Shruthi Sukumar, Adrian Ward, Camden Elliott-Williams, Shabnam Hakimi, Michael C. Mozer}

\ShortHeadings{Overcoming Temptation}{Sukumar, Ward, Elliott-Williams, Hakimi, Mozer}
\firstpageno{1}

\begin{document}

\title{Overcoming Temptation:\\ Incentive Design For Intertemporal Choice}
\author{
\name \hspace{-.06in}Shruthi Sukumar \email shruthi.sukumar@colorado.edu\\
\addr Department of Computer Science \\
University of Colorado at Boulder 
\AND
\name Adrian F. Ward 
\email adrian.ward@mccombs.utexas.edu \\
\addr McCombs School of Business \\
University of Texas at Austin 
\AND
\name Camden Elliott-Williams \email camden.elliottwilliams@colorado.edu\\
\addr Department of Computer Science \\
University of Colorado at Boulder
\AND
\name Shabnam Hakimi
\email shabnam.hakimi@duke.edu \\
\addr Center for Cognitive Neuroscience \\
Duke University
\AND
\name Michael C. Mozer \email mcmozer@google.com \\
\addr 
Google Research, Brain Team \\
and \\
Institute of Cognitive Science \\
University of Colorado, Boulder 
}

\maketitle
\newpage
\begin{abstract}
Individuals are often faced with temptations that can lead them astray from
long-term goals. We're interested in developing interventions that steer
individuals toward making good initial decisions and then maintaining those
decisions over time. In the realm of financial decision making, a particularly
successful approach is the prize-linked savings account: individuals are
incentivized to make deposits by tying deposits to a periodic lottery that
awards bonuses to the savers. Although these lotteries have been very
effective in motivating savers across the globe, they are a one-size-fits-all
solution. We investigate whether customized bonuses can be more effective. We
formalize a delayed-gratification task as a Markov decision problem and
characterize individuals as rational agents subject to temporal discounting,
a cost associated with effort, and fluctuations in willpower.
% willpower -> decision bias?
Our theory is able to explain key behavioral findings in intertemporal choice.
We created an online delayed-gratification game in which the player scores
points by selecting a queue to wait in and then performing a series of actions to 
advance to the front. Data collected from the game is fit to the model, and the 
instantiated model is then used to optimize predicted player performance over a 
space of incentives. We demonstrate that customized incentive structures can
improve an individual's goal-directed decision making.
\end{abstract}

\vspace{1in}

\begin{significancestatement}
Individuals are often tempted to abandon long-term goals (e.g., weight loss) by enticements that provide immediate reward (e.g., a piece of cake). We use computational models of decision making to determine personalized interventions that allow an individual to overcome temptation and improve long-term outcomes. We formalize a theory in which individuals make a series of choices to \emph{persist} toward long-term goals or \emph{defect} and obtain an immediate reward. The theory is used to determine a limited schedule of incentives that maximizes expected outcomes.  We conduct experiments with a simulated line-waiting task that show the theory's potential.
\end{significancestatement}
\newpage

Should you go hiking today or work on that manuscript? Should you have a slice
of cake or stick to your diet? Should you upgrade your flat-screen TV or 
contribute to your retirement account? Individuals are regularly faced
with temptations that lead them astray from long-term goals. These temptations
all reflect an underlying challenge in behavioral control that involves
choosing between actions leading to small but immediate rewards and actions
leading to large but delayed rewards. We introduce
a formal model of {\em delayed-gratification} decision tasks
% REMOVED BY MIKE 2/13/2022
%extending the Markov decision framework to incorporate psychological notions
%of willpower and effort, and using formal models 
and use the model
to optimize behavior by designing
incentives to assist individuals in achieving long-term goals.

Consider the serious predicament with retirement planning in the United States.
Only 55\% of working-age households have retirement account
assets---whether an employer-sponsored plan or an IRA---and the median account
balance for near-retirement households is \$14,500. Even considering
households' net worth, $2/3$ fall short of conservative savings targets based
on age and income \citep{RheeBoivie2015}. 
Balances in retirement accounts for age-60 participants are reduced
by 31\% due to leakage, including cash-outs, hardship withdrawals, and
the failure to repay loans \citep{Goodman2021}.
%Furthermore, 40\% of every dollar
%contributed to the accounts of savers under age 55 simultaneously flows out of
%the retirement systems, not counting loans to oneself
%\citep{ArgentoBryantSabelhaus2015}. 
In 2013, the US government and nonprofits
spent \$670M on financial education, yet financial literacy accounts for a
minuscule 0.1\% of the variance in financial outcomes
\citep{FernandesLynchNetemeyer2014}.

One technique that has been extremely successful in encouraging savings,
primarily in Europe and the developing world but more recently in the US as
well, is the {\em prize linked savings account (PLSA)}
\citep{KearneyTufanoGuryanHurst2010, gertler2019increasing}. The idea is to pool a fraction of the
interest from all depositors to fund a prize awarded by periodic lotteries.
Just as ordinary lotteries entice individuals to purchase tickets, the PLSA
encourages individuals to save. Disregarding the fact that lotteries function
in part because individuals overvalue low-probability gains
\citep{KahnemanTversky1979}, the core of the approach is to offer savers the
prospect of short-term payoffs in exchange for them committing to the long
term. Although the account yields a lower interest rate to fund the lottery,
the PLSA increases the net expected account balance due to
greater commitment to participation.

The PLSA is a one-size-fits-all solution. A set of incentives that
work well for one individual or one subpopulation may not be optimal for 
another. In this article, we investigate approaches to customizing incentives to 
an individual or a subpopulation with the aim of achieving greater adherence to
long-term goals and ultimately, better long-term outcomes for the participants.
Our approach involves: (1) building a model to characterize the behavior of
an individual or group, (2) fitting the model with behavioral data, 
(3) using the model to determine an incentive structure that optimizes 
outcomes, and (4) validating the model by showing better outcomes with
model-derived incentives than with alternative incentive structures.

\section*{Intertemporal Choice}
Saving for retirement and other delayed-gratification tasks involve 
choosing between alternatives that produce gains and losses at 
different points in time, or \emph{intertemporal choice}.
How an individual interprets delayed consequences
influences the utility or value associated with a decision. When consequences
are discounted with the passage of time, decision making leans toward more
immediate gains and more distant losses. The {\em delay discounting} task is
often used to study intertemporal choice \citep{GreenMyerson2004}. Individuals
are asked to choose between two alternatives, e.g., \$1 today versus \$$X$ in
$Y$ days. By identifying the $X$ that yields subjective indifference for a
given $Y$, one can estimate an individual's discounting of future outcomes.
Discount rates vary across individuals yet show stability over extended periods
of time \citep{Kirby2009}.

%from McGuire paper:
%temporal discount rate, which varies across individuals and shows
%trait-like stability over time (Kirby, 2009; Ohmura, Takahashi, Kitamura, \&
%Wehr, 2006).

% Adrian's distinction between initial decision and decision maintenance
%\color{blue}
This paradigm involves a single, hypothetical decision and reveals the
intrinsic future value of an outcome. However, it does not address the temporal
dynamics of behavior over an extended period of time in delayed-gratification tasks.
In such tasks, once an initial decision is made
to wait for a large reward, individuals are permitted to abandon the
decision \textit{at any instant} in favor of the small immediate reward.
For example, in the classic marshmallow test \citep{MischelEbbesen1970},
children are seated at a table with a single marshmallow. They are allowed to
eat the marshmallow, but if they wait while the experimenter steps out of the
room, they will be offered a second marshmallow when the experimenter returns.
In this scenario, children continually contemplate
whether to eat the marshmallow or wait for two marshmallows. Their
behavior depends not only on the hypothetical discounting of future rewards but
on the individual's {\em willpower} \cite{Ainslie2021}---their ability to maintain focus on the
larger reward and not succumb to temptation before the experimenter returns.
Defection at any moment eliminates the possibility of the larger reward.

The marshmallow test achieved renown not only because it is claimed to be
predictive of later life outcomes \citep{MischelShodaRodriguez1989}, but
because it is analogous to many situations involving delayed gratification.
Like the marshmallow test, some of these situations have an unspecified time
horizon (e.g., exercise, waiting for an elevator, spending during retirement).
However, others have a known horizon (e.g., avoiding snacks before dinner,
saving for retirement, completing a college degree). Our work addresses the
case of a known or assumed horizon.
%Another timed example: college planning (take fun courses
%versus courses needed for graduation).
%Duration is known:
%When one is 50, one has a much greater interest in retirement issues than
%when one is 20. 

Whether or not the horizon is known, delayed-gratification tasks may
additionally be characterized in terms of the number of opportunities to obtain the
delayed reward. The marshmallow test is \textit{one shot}, but many
true-to-life scenarios have an \textit{iterated} nature. For example, in
retirement planning, the failure to contribute to the account one month does
not preclude contributing the next month. Another intuitive example involves
allocating time within a work day. One must choose between tasks that are
brief and provide a moment of satisfaction (e.g., answering email)
and tasks that will take a long time to complete but will eventually
yield a sense of accomplishment (e.g., writing a manuscript). Our 
work addresses both one-shot and iterated 
delayed-gratification tasks. For such tasks, we're interested in developing
personalized interventions that assist individuals both in making good initial
decisions and in maintaining those decisions over time.
\color{black}

%\color{blue}
\section*{Theories of Intertemporal Choice}

Nearly all previous conceptualizations of intertemporal choice have focused on
the shape of the discounting function and the initial `now versus later'
decision, not the time course. One exception is the work of
\citet{McGuireKable2013} who frame failure to postpone gratification as a
rational, utility-maximizing strategy when the time at which future outcomes
materialize is uncertain. Our theory is complementary in providing a rational
account in the known time horizon situation.

When one considers the time course of delaying gratification---the need to recommit
to the decision and not succumb to temptation---the appropriate framework for
modeling behavior is that of sequential decision making.
There is a rich literature on modeling human sequential decision-making using the
formalism of Markov decision processes \citep[\textit{MDP}s; 
e.g.,][]{niv2007tonic,NivEdlundDayanODoherty2012,ShenTobiaSommerObermayer2014,Lieder2019tutor,Lieder2019, Drummond2020,Widrich2021,Bastani2021}. 
In this framework, human behavior is interpreted with respect to the behavior 
of a rational agent, i.e., an agent following an optimal (reward maximizing) policy.
The optimal policy is determined via dynamic programming or reinforcement learning
\citep{SuttonBarto2018}, both of which can accommodate discounting of future outcomes.
Even when human decisions deviate from those of the rational agent, the modeling framework is
nonetheless valuable if additional \emph{bounded rationality} assumptions allow it to
account for human performance \citep{Simon1997,ToddGigerenzer2012,ChaterOaksford1999}.

The MDP framework, which allows rewards of various magnitudes to be realized at different points in time,
is well suited for modeling intertemporal choice. 
\citet{KurthNelson2010,KurthNelson2012} explore a model of precommitment in decision making as a means of preventing impulsive defections. Their model addresses the initial decision to commit rather than the ongoing possibility of defection. \citet{Lieder2019}, whose work is closest to our own, address a class of intertemporal-choice tasks---not exactly delayed gratification tasks---in which individuals have to choose between difficult and time consuming work assignments that eventually lead to a large payout (strategic route planning, essay writing) and easy alternatives that thwart obtaining the eventual payout (making an impulsive choice, watching a YouTube video, abandoning the experiment). They use the MDP theoretical framework to
develop an \emph{optimal gamification} approach to help individuals avoid procrastination and to achieve
future-minded goals. We further discuss this interesting work and its relation to ours later and in the Supplementary Materials.

%In recent research most closely related to ours, \citet{Lieder2019} consider tasks that require individuals to choose between high-paying work assignments that are difficult and effortful (synthetic route planning, essay writing) and lower-paying alternatives that are easy or can be achieved with shortcut heuristics. Although they consider tasks in which people generally  procrastinate and fail to achieve future-minded goals, they do not use the MDP framework to  model human behavioral data directly. We return to their work shortly.

\section*{Formalizing Delayed-Gratification Tasks as a Markov Decision Problem}

In this section, we formalize a delayed-gratification task as a Markov
decision problem, which we will refer to as the {\em DGMDP}. 
We assume time to be quantized into discrete
steps and we focus on situations with a known or assumed time horizon, denoted
$\tau$. At any 
step, the agent may {\em \defect} and collect a small reward, or the agent 
may {\em \persist} to the next step, eventually collecting a large reward
at step~$\tau$. We use $\rss$ and $\rll$ to denote the 
{\em smaller sooner (SS)} and {\em larger later (LL)} rewards.
%\begin{figure}[bt]%[100]%[bt]
%  \includegraphics[width=6in]{fig/FSMabcd.pdf}
%  \caption{Finite-state environment formalizing (a) the one-shot 
%  delayed-gratification task; (b) the iterated 
%  delayed-gratification task; (c) the iterated delayed-gratification task
%  with variable delays and LL outcomes; and (d) an efficient approximation
%  to the iterated delayed-gratification task, suitable when episodes are
%  independent of one another.
%  \label{fig:FSM}}
%\end{figure}
\begin{figure}[bt]%[100]%[bt]
  \includegraphics[width=6in]{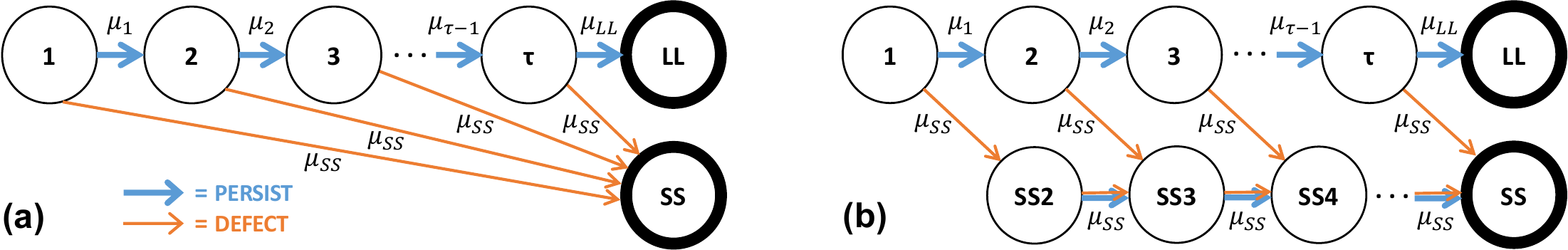}
  \caption{Finite-state environment formalizing (a) the one-shot 
    delayed-gratification task; (b)  an efficient approximation
    to the iterated delayed-gratification task, suitable when episodes are
    independent of one another and reward rate is to be maximized.}
  \label{fig:FSMmain}
\end{figure}
Figure~\ref{fig:FSMmain}a shows a finite-state representation of the one-shot task
with terminal states LL and SS that correspond to
resisting and succumbing to temptation, respectively, and states for each
time step between the initial and final times, $t \in \{1, 2, ..., \tau\}$. 
Rewards are associated with state transitions. The possibility of obtaining 
{\em intermediate} rewards during the delay period is annotated via
$\ri \equiv \{ \mu_1, ..., \mu_{\tau-1} \}$, which we return to later.
With exponential discounting,
rewards $n$ steps ahead are devalued by a factor of 
$\gamma^n$, $0 \le \gamma < 1$. 

Given the DGMDP, an optimal decision 
sequence is trivially obtained by value iteration. However,
this sequence is a poor characterization of human behavior. With no intermediate
rewards ($\smash{\ri = \mathbf{0}}$), it takes one of two
forms: either the agent defects at $t=1$ or the agent persists through
$t=\tau$. In contrast, individuals will often persist some time and then
defect, and when placed into the same situation repeatedly, behavior
is nondeterministic. For example, replicability on the marshmallow
test is quite modest, with $\rho<0.30$ \citep{MischelShodaPeake1988}.
The discrepancy between human delayed-gratification behavior and the optimal 
decision-making framework might indicate an incompatibility. However, we 
prefer a bounded-rationality perspective on human cognition according to 
which behavior is cast as optimal but subject to operational constraints \citep{Simon1997}.
We postulate two specific constraints. 
\setdefaultleftmargin{12pt}{0pt}{}{}{}{}
\begin{enumerate}
\item
The decision-making framework is one component of a cognitive architecture. When modeling an isolated
component, it is common to treat factors external to the component as a noise source that contributes
to behavioral variability. Here, we introduce a one-dimensional Gaussian process, $W = \{ W_t \}$, with 
\[
w_1~\sim~\mathrm{Gaussian}(0, \sigma_1^2) \qquad\text{and}\qquad
w_t~\sim~\mathrm{Gaussian}(w_{t-1}, \sigma^2) .
\]
We suppose that this quantity, which we refer to as the \emph{\bias}, modulates an individual's subjective value of defecting at step $t$:
\begin{equation}
Q(\{t, w\}, \defect) = \rss - w,
\label{eq:qdefect}
\end{equation}
where $Q(\boldsymbol{s},a)$ 
denotes the value associated with performing action $a$ in state $\boldsymbol{s}$,
and the state space consists of the discrete step $t$ and the continuous 
\bias $w$.\footnote{For further discussion and justification of this  assumption, please see the Supplementary Materials.}
%We chose the variable $w$ because the process represents something vaguely like fluctuations in willpower. 
%We opted for an additive bias mainly for reasons of
%mathematical convenience.
\item
%The second constraint that we include in the model is motivated by
Behavioral, economic, and neural accounts of decision making suggest that
{\em effort} carries a cost, that rewards are weighed against the effort
required to obtain it \citep[e.g.,][]{Kivetz2003}, and that the avoidance of
effort has mechanistic and rational bases \citep{shenhavetal2017}. 
Without concerning ourselves with these bases, we incorporate into the
model an effort cost, $\re$, that is associated with persevering:
\begin{equation}
%Q(t,w_t;\text{\persist}) = \re + \mu_t + \gamma 
%\mathbb{E}_{w_{t+1}|w_t} [ \mathrm{max}_a Q(t+1,w_{t+1};a) ] 
Q(\{t,w\}, \persist) = 
\begin{cases}
\re + \mu_t + \gamma ~
	 \mathbb{E}_{W_{t+1}|W_t=w} V(t+1, w_{t+1}) & \text{for }t < \tau \\
\rll & \text{for }t = \tau
\end{cases}
\label{eq:qpersist}
\end{equation}
\begin{equation}
\text{where}~~ V(t, w) \equiv \mathrm{max}_a ~Q(\{t,w\},a) .
%\text{where}~~ V(t+1, w_{t+1}) \equiv \mathrm{max}_a Q(t+1,w_{t+1};a) .
\label{eq:value}
\end{equation}
\end{enumerate}
With these two constraints, we will show that the model not only has adequate
expressive power to fit behavioral data, but also has the explanatory power
to predict experimental outcomes.

The one-shot DGMDP in Figure~\ref{fig:FSMmain}a can be extended to model the
iterated task, even when there is variability in the reward ($\rll$) or duration ($\tau$) 
across \textit{episodes} (see Figure~Supp-\ref{fig:FSMsupp}a,b). It is straightforward 
to show that the solution to the iterated DGMDPs is identical to the solution to the 
simpler and  more tractable one-shot DGMDP in Figure~\ref{fig:FSMmain}b under certain 
constraints  (see Supplementary Information). Essentially, Figure~\ref{fig:FSMmain}b 
models the  choice between the LL reward or a sequence of SS rewards matched in total 
number of  steps, effectively comparing the reward rates for LL and SS, the critical 
variables in the iterated DGMDP.

%The one-shot DGMDP in Figure~\ref{fig:FSMmain}a can be extended to model the
%iterated task (Figure~\ref{fig:FSM}b), even when there is variability in the
%reward ($\rll$) or duration ($\tau$) across \textit{episodes}
%(Figure~\ref{fig:FSM}c).\footnote{Figures~\ref{fig:FSM}b,c describe an
%indefinite series of episodes. If the total number of episodes or steps is
%constrained, as in any realistic scenario (e.g., an individual has eight hours
%in the work day to perform tasks like answering email), then the state must be
%augmented with a representation of remaining time. We dodge this complication
%by modeling situations in which the `end game' is not approaching, e.g., only
%the first half of a work day.} Finally, it is straightforward to show that the
%solution to the iterated DGMDPs in Figures~\ref{fig:FSM}b or \ref{fig:FSM}c is
%identical to the solution to the simpler and more tractable one-shot DGMDP in
%Figure~\ref{fig:FSM}d under certain constraints (see Supplementary Information).
%Essentially, Figure~\ref{fig:FSM}d models the choice between the LL reward or a
%sequence of SS rewards matched in total number of steps, effectively comparing
%the reward rates for LL and SS, the critical variables in the iterated DGMDP.

To summarize, we have formalized one-shot and iterated delayed-gratification
task with known horizon as a Markov decision problem with parameters
$\Theta_{\mathrm{task}} \equiv \{ \tau, \rss, \rll, \ri \}$, and a constrained
rational agent parameterized by $\Theta_{\mathrm{agent}} \equiv \{ \gamma,
\sigma_1, \sigma, \re \}$. We now turn to solving the DGMDP and characterizing
its properties.

\subsection*{Solving The Delayed-Gratification Markov Decision Problem (DGMDP)}

\newcommand{\at}{\ensuremath{a_{t}}}
\newcommand{\bt}{\ensuremath{b_{t}}}
\newcommand{\ct}{\ensuremath{c_{t}}}
\newcommand{\zmt}{\ensuremath{z_t^{-}}}
\newcommand{\zpt}{\ensuremath{z_t^{+}}}
The simple structure of the environment allows for a backward-induction
solution to the Bellman equation (Equation~\ref{eq:qpersist}). Although the \bias $w$ precludes an analytical solution for the value
$V(t,w)$, we construct a piecewise-linear approximation over 
$w$ for each step $t$, as described in the Supplementary Materials.
Figure~\ref{fig:value}a shows the value as a function of \bias at each step of an eight
step DGMDP with an LL reward twice that of the SS reward, like the canonical
marshmallow test. Both the exact value-function formulation obtained by 
discretizing \bias and the corresponding piecewise-linear approximation 
(Equation~Supp-\ref{eq:valueinduction}) 
are presented in colored and black lines, respectively. 

\begin{figure}[tbp]
  \centering
  \includegraphics[width=5.00in]{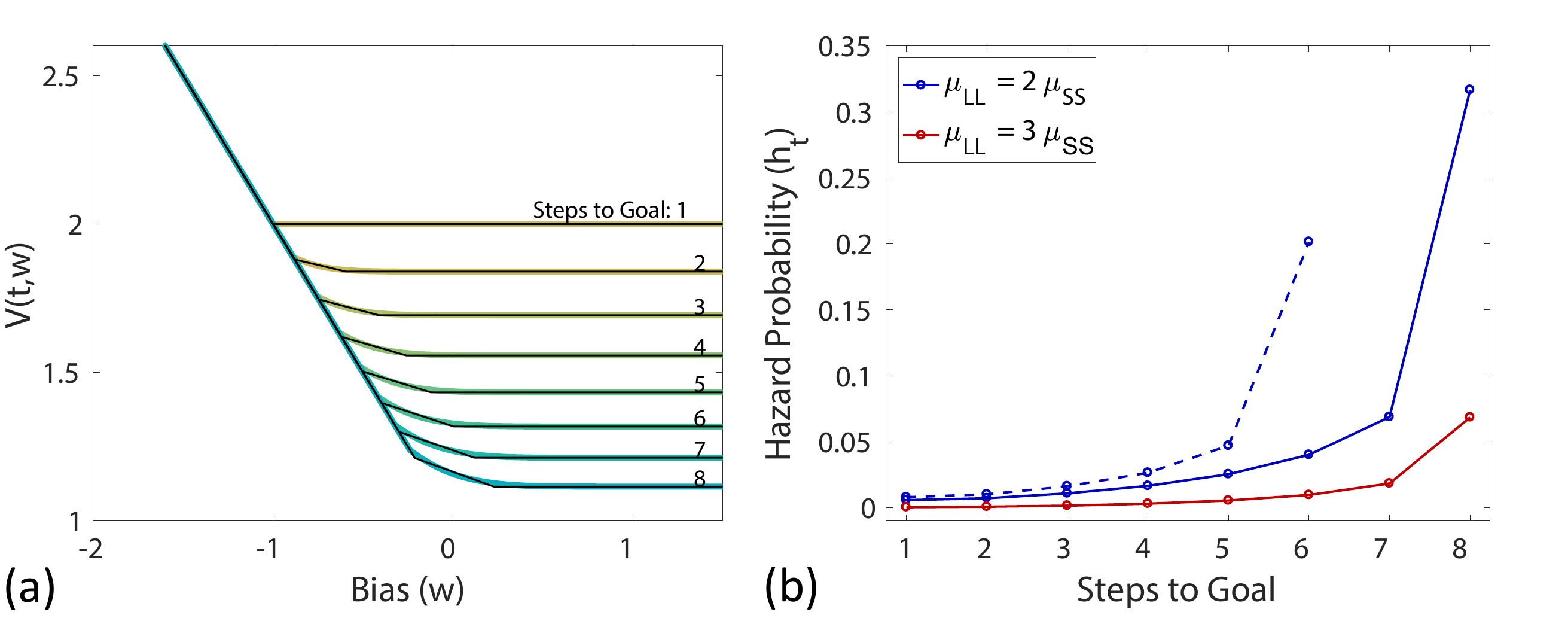}
  \caption{(a) Value function for a DGMDP with $\tau=8$,
  $\sigma=.25$, $\sigma_1 = .50$, $\gamma=.92$, $\re=\mu_t=0$, $\rll=2$, $\rss=1$,
  exact (colored curves) and piecewise linear approximation (black lines).
  (b) Hazard functions for the parameterization in (a) (solid blue curve),
  with a higher level of LL reward (red curve), and with a shorter delay
  period, $\tau=6$ (dashed blue curve).
  \label{fig:value}}
\end{figure}

Using the value function, we can characterize the agent's behavior in
the DGMDP via the likelihood of defecting at various steps. With $D$ denoting
the defection step, we have the {\em hazard probability},
\begin{equation}
h_t \equiv P(D = t | D \ge t) 
\equiv P(W_t < w_t^* | W_1 \ge w_1^*, ..., W_{t-1} \ge w_{t-1}^*),
\label{eq:hazard}
\end{equation}
where
$w^*$ is the \bias threshold that yields action indifference,
\begin{equation*}
Q(t,w^*;\defect) = Q(t,w^*;\persist) . 
\end{equation*}
Estimation of Equation~\ref{eq:hazard}
is discussed in the Supplementary Materials.

The solid blue curve in Figure~\ref{fig:value}b shows the hazard function for 
the DGMDP in Figure~\ref{fig:value}a. Defection rates drop as
the agent approaches the goal. Defection rates also scale with the LL reward,
as illustrated by the contrast between the solid blue and red curves.
Finally, defection rates depend both on relative and absolute steps to goal:
contrasting the solid and dashed blue curves, corresponding to $\tau=8$ and
$\tau=6$, respectively, the defection rate at a given number of steps
from the goal depends on $\tau$. We show shortly that human 
data exhibit this same qualitative property. Interestingly, the random walk 
in \bias is critical in obtaining this
property. When \bias is independent from step to step, i.e., 
$w_t~\sim~\mathrm{Gaussian}(0, \sigma^2)$, defection rates depend only on
absolute steps to goal. Thus, the moment-to-moment nonzero autocorrelation
is essential for modeling human behavior.

\subsection*{Behavioral Phenomena Explained}

We consider the solution of the DGMDP as a rational theory of human
cognition. It is meant to explain both an individual's initial choice (``Should
I open a retirement account?'') as well as the temporal dynamics of 
sustaining that choice
%individual's adherence to that choice 
(``Should I withdraw the funds to buy a car?'').

Our theory explains two key phenomena in the literature. First,
failure on a DG task is sensitive to the relative magnitudes of the SS and LL
rewards \citep{Mischel1974}. Figure~\ref{fig:value}b presents hazard
functions for two reward magnitudes. The probability of obtaining the LL
reward is greater with $\rll/\rss = 3$ than with $\rll/\rss = 2$. 
Figure~\ref{fig:value}b can also accommodate the finding that environmental 
reliability and trust in the experimenter affect outcomes in the marshmallow 
test \citep{KiddPalmeriAslin2012}: in unreliable or nonstationary environments, 
the expected LL reward is lower than the advertised reward, and the DGMDP
is based on reward expectations. Second, a reanalysis of data from
a population of children performing the marshmallow task shows a declining
hazard rate over the task period of 7 minutes \citep{McGuireKable2013}. 
The rapid initial drop in the
empirical curve looks remarkably like the curves in Figure~\ref{fig:value}b.
One might interpret this phenomenon as a {\em finish-line effect}: the closer
one gets to a goal, the greater is the commitment to achieve the goal.
However, the model suggests that this behavior arises not from abstract
psychological constructs but because of correlations in \bias over time:
if an individual starts down the path to an LL reward, the individual's
\bias at that point must be high. The posterior \bias distributions
reflect the elimination of individuals with low momentary \bias, which
contributes to the declining hazard rate. Also contributing is the exponential
increase in value of the discounted LL reward as the agent advances through
the DGMDP. \citet{McGuireKable2013} explain the empirical hazard function via
a combination of uncertainty in the time horizon and time-fluctuating 
discount rates. Our theory shows that these strong assumptions are not 
necessary, and our theory can address situations with a well delineated
horizon such as retirement saving. Additionally, our theory aims to move
beyond population data and explain the granular dynamical behavior of an 
individual, as we demonstrate in experiments to follow. 
\section*{Optimizing Incentives}
We explore a mechanism-design approach \citep{NisanRonen1999} aimed at steering  individuals toward improved long-term outcomes. We ask whether we can provide incentives or \emph{bonuses} to rational value-maximizing agents that will increase their expected reward. In contrast to \citep{Lieder2019}, the bonuses are actually paid out and are constrained so as not to ``print money,'' as we describe shortly.

We first address an investment scenario roughly analogous to a prize-linked savings
account (PLSA). Suppose an individual has $x$ dollars which they can deposit
into a bank account earning interest at rate $r$, compounded annually. At the
start of each year, they decide whether to continue saving (\persist) or to
withdraw and spend their {\em entire} savings with interest accumulated thus
far (\defect).\footnote{Although this all-or-none withdrawal of savings is not
entirely realistic, it reduces the decision space to correspond with the FSM in
Figure~\ref{fig:FSMmain}a. Were we to allow intermediate levels of withdrawal, the
simulation would yield intermediate benefits of incentives.} Our goal to assist
them in maximizing the profit they reap over $\tau-1$ years from their initial
investment. Our incentive mechanism is a a schedule of lotteries. We refer to
expected lottery distributions as {\em bonuses}, even though they are funded
through the interest earned by a population of individuals, like the prizes of
the PLSA.

With $\mu_t$ denoting the bonus awarded in year $t$ and $\ri$ denoting the set
of scheduled bonuses, our goal as mechanism designers is to identify the
schedule that maximizes the expected net accumulation from an individual's
investment: 
\begin{equation} 
\textstyle{\bm{\mu}^*_{1:\tau-1} =
\text{argmax}_{\ri} \sum_{t=1}^{\tau} P(D=t | \gamma, \ri ) \left[ b_t +
\sum_{t'=1}^{t-1} \mu_{t'} \right]} , 
\end{equation} 
where $b_t$ is the amount banked at the start of year $t$, with $b_1 = x$ and
$b_{t+1} = (1+r) (b_t - \mu_t)$, and $D$ is the year of defection, where $D=1$
represents immediate defection and $D=\tau$ represents the the account reaching
maturity. Defection probabilities are obtained from the theory
(Equation~\ref{eq:hazard}).

\begin{figure}[btp]
  \centering
  \includegraphics[width=6.00in]{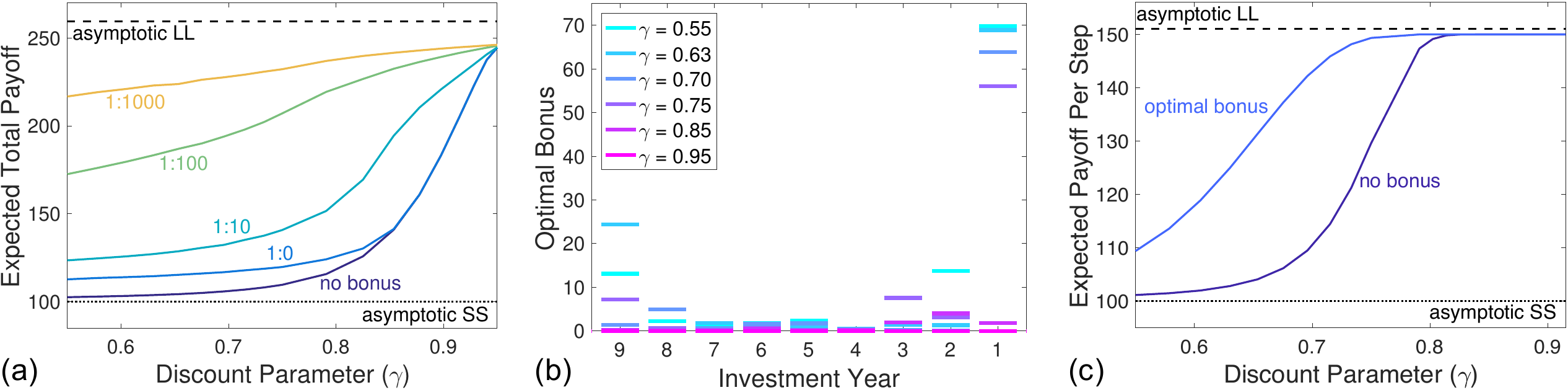}
  \caption{Bonus optimization for an agent with $\sigma_1=50$, $\sigma=30$,
  $\re=0$, and $\gamma \in [0.55, 0.95]$. (a) Expected payoff for the
  one-shot DGMDP for various bonus scenarios, including no bonus and optimal
  bonuses with lottery odds 1:0, 1:10, 1:100, and 1:1000. In these simulations,
  the interest-accrual setting is used to constrain bonuses and payoffs.
  (b) For the certain win (1:0) lottery with an initial fund of $x=100$,
  optimal bonus at each step for various $\gamma$. The optimal bonus is expressed
  as the percentage of initial pool of funding. (c) Expected payoff per time
  step for the iterated DGMDP for the bonus-limit setting used to constrain
  bonuses and payoffs.}
  \label{fig:optbonus}
\end{figure}

To illustrate this approach, we conducted a simulation with discount factor
$\smash{\gamma \in [0.55,0.95]}$, $\smash{\tau=10}$ year horizon, annual
interest rate $\smash{r=0.1}$, and initial bank $\smash{x=100}$,
comparing an agent's expected accumulation without bonuses and with optimal
bonuses. Optimization is via direct search using the simplex algorithm over
unconstrained variables $p_t \equiv \text{logit} (\mu_t / b_t)$, representing
the proportion of the bank being distributed as a bonus. 

We first consider the case of deterministic bonuses: the agent receives bonus
$\mu_t$ in year $t$ with certainty. Figure~\ref{fig:optbonus}a shows the
expected payoff as a function of an agent's discount factor $\gamma$ for the
scenario with no bonuses (purple curve) versus optimal bonuses awarded with
probability 1.0 (light blue curve, labeled with the odds of a bonus being
awarded, `1:0'). For reference, the asymptotic SS and LL payoffs are shown with
dotted and dashed lines, respectively.

With high discounting, this simulation yields a roughly $10\%$ improvement
in an individual's expected accumulation by providing bonuses at the end of the
early years and going into the final year (Figure~\ref{fig:optbonus}b).
Bonuses are recommended only when the gain from encouraging persistence beats
the loss of interest on an awarded bonus. With low discounting, the model
optimization recommends no bonuses. Thus, the simulation recommends different
incentives to individuals depending on their discount factors.
%\mcm{
%Trivially, the benefit of a bonus increases when coupled with a penalty for
%early withdrawals---another aspect of mechanism design (not shown in Figure).
%}

Now consider a lottery such as that conducted for the PLSA. If individuals
operate based on expected returns, an uncertain lottery with odds 1:$\alpha$
and payoff $(\alpha+1)\mu_t$ would be equivalent to a certain payoff of
$\mu_t$. However, as characterized by prospect theory
\citep{KahnemanTversky1979}, individuals overweight low probability events.
Using median parameter estimates from cumulative prospect theory
\citep{TverskyKahneman1992} to infer subjective probabilities on lotteries with
1:10, 1:100, and 1:000 odds, we optimize bonuses for these
cases.\footnote{According to prospect theory, the 1:10, 1:100, and 1:1000
lotteries yield overweighting by factors of 1.86, 5.50, and 14.40,
respectively.} As depicted by the three upper curves in
Figure~\ref{fig:optbonus}a, lotteries such as the PLSA can significantly boost
the benefit of incentive optimization.

Lotteries and interest accrual are not suitable for all delayed-gratification
tasks. For instance, one would not wish to encourage a dieter by offering a
lottery for a 50-gallon tub of ice cream or the promise of a massive
all-one-can-eat dessert buffet at the conclusion of the diet. To demonstrate
the flexibility of our framework, we posit a \textit{bonus-limit} setting as an
alternative to the \textit{interest-accrual} setting in which up to $n_b$
bonuses of fixed size can be awarded and the optimization determines the time
steps at which they are awarded. We conducted a simulation with the iterated
DGMDP (Figure~\ref{fig:FSMmain}b) using $\smash{\gamma \in [0.55,0.95]}$,
$\smash{\tau=10}$, awarding of $n_b \le 4$ bonuses each of value 50,
$\smash{\rss = 100}$, and $\smash{\rll = 150\tau-50n_b}$. Multiple bonuses
could be awarded in the same step, but bonuses were limited such that no
defection could achieve a reward rate greater than $\smash{\rss}$. This setting
ensures that bonuses
corresponds to the conditions used for human experiments that we report on next.
Figure~\ref{fig:optbonus}c shows expected payoff per step, ranging from 100
from the SS reward to 150 for the LL reward, for the no-bonus and optimal-bonus
conditions. As with the alternative DGMDP formulation with a single-shot
task and the interest-accrual setting, optimization of bonuses in the bonus-limit 
setting yields bonus distributions and benefits that depend on discount factor $\gamma$.

\section*{Human Behavioral Experiments}
%%%%%%%%%%%%%%%%%%%%%%%%%%%%%%%%%%%%%%%%%%%%%%%%%%%%%%%%
% REMOVED FOR LENGTH
%Our modeling framework is flexible enough to describe a variety of delayed gratification tasks, both one shot and iterative, with variable payoff and incentive structures. This framework provides a potential explanation of human cognition, under the conjecture that individuals can be cast as bounded rational agents who seek to maximize their payoffs given cognitive constraints such as discounting and fluctuations in willpower. If this conjecture is supported, the framework should allow us to determine incentives that will shape behavioral outcomes.
%
%Typically, support for a model is obtained by comparing it to alternatives and arguing that one model is better on grounds of parsimony or predictive power. With no existing models suited to explaining the moment-to-moment dynamics of behavior, our strategy instead is to show first that the model is consistent with behavior by fitting model parameters to behavioral data (Experiments 1-3), and second, that the fitted, fully constrained model can make strong predictions concerning the conditions under which an individual's behavior will be optimized for a targeted long-term outcome (Experiments 4 and 5).

\begin{figure}[btp]
  \centering
  \includegraphics[width=6in]{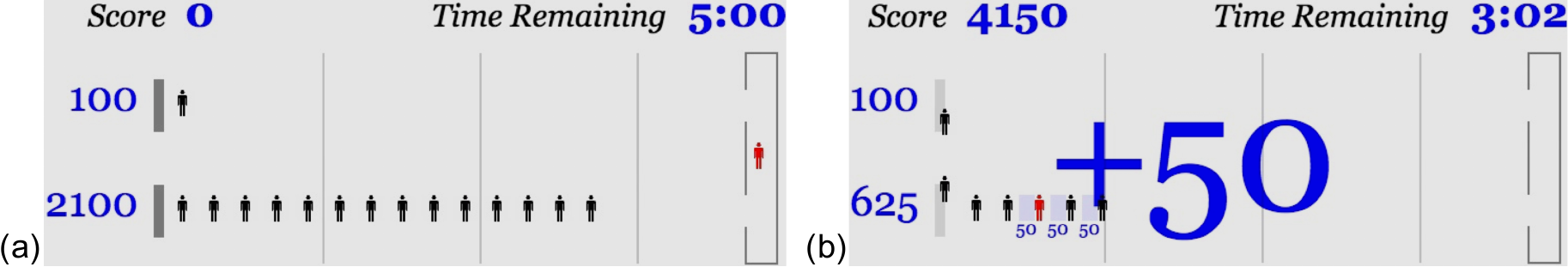}
  \caption{The queue-waiting game. (a) The player (red icon) is in the 
  vestibule, prior to choosing a queue. Queues advance right to left. 
  Points awarded per queue are displayed left of the queue.
  (b) A snapshot of the game taken while
  the queues advance. As described in the text, this condition 
  includes bonuses at certain positions in the long queue. Point increments
  are flashed as they are awarded.}
  \label{fig:queuewaiting}
\end{figure}

%We created an online delayed-gratification game in which players
%%%%%%%%%%%%%%%%%%%%%%%%%%%%%%%%%%%%%%%%%%%%%%%%%%%%%%%%
To evaluate the model's ability to recommend bonuses that improve long-term outcomes,
we created an online delayed-gratification game in which players
score points by waiting in a queue, much as diners load their plates with food
by waiting their turn at a pre-pandemic restaurant buffet 
(Figure~\ref{fig:queuewaiting}a). The upper queue is short, having only one
position, and the lower queue is long, having $\tau$ positions. The 
minimum time to obtain a reward has a ratio of $\tau : 1$ for the
long versus short queues. When the player is serviced, the short and long queues deliver 
a $100$ and $100\tau\rho$ point reward, respectively. The {\em reward-rate ratio}, 
$\rho$, is either 1.25 or 1.50 in our experiments.  The player starts in a
vestibule (right side of screen) and selects a queue with the up and down arrow keys.
The game updates at a fixed interval (1000 or 2000 msec), at which point the player's request is
processed and the queues advance (from right to left). Upon entering the short
queue, the player is immediately serviced. Upon entering the long queue, the
player immediately advances to the next-to-last position as the queue shuffles
forward. With every tick of the game clock, the player may hit the left-arrow
key to advance in the long queue or the up-arrow key to defect to the short
queue. If the player takes no action, the simulated participants behind the
player jump past. When the player defects to the short queue, the player is
immediately serviced. When points are awarded, the screen flashes the points
and a cash register sound is played, and the player returns to the vestibule
and a new {\em episode} begins. For each episode, a long queue length $\tau$ 
is drawn randomly, with lengths ranging from 4 to 14.

Note that the reward rate (points per action) for either
queue does not depend on the long-queue length. Because of this constraint,
each episode is functionally decoupled from following episodes. That is, the
optimal action for the current episode will not depend on upcoming 
episodes.\footnote{A dependence does occur in the final seconds of the
game, where the player may not have sufficient time to complete the long 
queue. We handle this case by showing the player the time remaining, and discarding
game play in the last 30 seconds of the game.}
Due to this fact and the time-constrained nature of the game, the iterated
DGMDP in Figure~\ref{fig:FSMmain}b is appropriate for describing a rational player's
understanding of the game. This DGMDP focuses on reward \textit{rate} and
treats a defection as if the player continues to defect until $\tau$ steps are
reached, each step delivering the small reward. 
The vestibule in Figure~\ref{fig:queuewaiting}a corresponds to state 1 in 
Figure~\ref{fig:FSMmain}b and lower queue position closest to the service desk 
to state $\tau$. Note the left-to-right reversal of the two Figures, which has 
often confused the authors of this article.

%If one long-queue had a better return than another long-queue per timestep, 
%the optimal strategy might be to defect in order to advance to the next 
%episode. If time was running out, the optimal strategy may be to defect
%to harvest reward before the timer runs out.

\subsection*{Experiment 1: Varying Reward Magnitude}

\begin{figure}[tb]%[100]%[bt]
  \centering
  \includegraphics[width=4in]{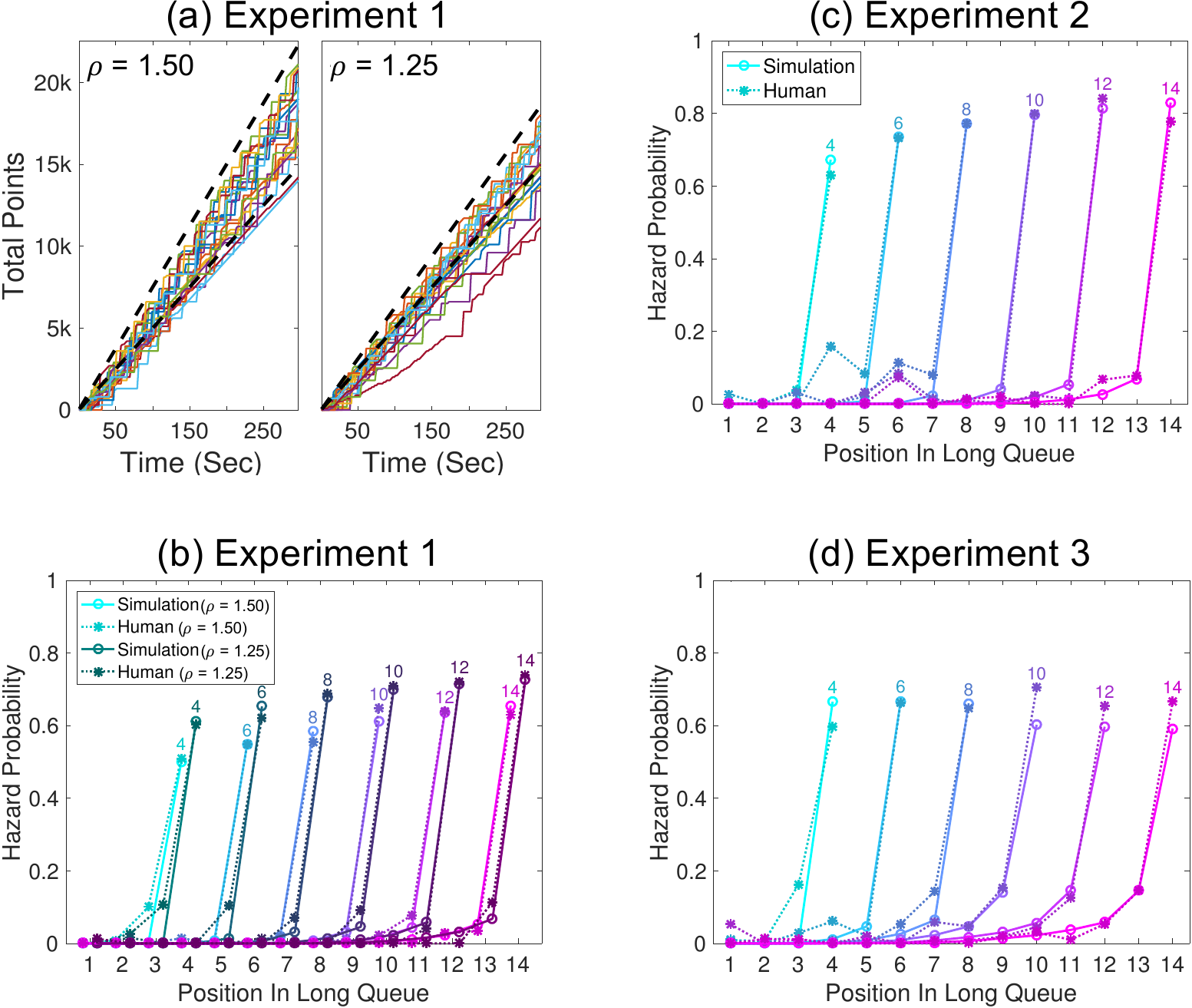}
  \caption{(a) Game points accumulated by individual participants over time
  in Experiment 1. (b) Hazard curves in Experiment 1 for 6 line lengths and
  two reward-rate ratios. Human data shown with asterisks and dashed lines, model
  fits with circles and solid lines. (c) Hazard curves for Experiment 2,
  with only one free model parameter, (d) Hazard curves for Experiment 3,
  with no free model parameters.}
  \label{fig:results1}
\end{figure}

Experiment 1 tested reward-rate ratios ($\rho$) 1.25 and 1.50.
Figure~\ref{fig:results1}a shows the reward
accumulation by individual participants in the two conditions as a function of
time within the session. The two dashed black lines represent the reward that
would be obtained by deterministically performing the SS or LL action at each
tick of the game clock. (Participants are not required to act every tick.) The
traces show that some participants had a strong preference for the short queue,
others had a nearly perfect preference for the long queue, and still others
alternated between strategies. The variability in strategy over time within
an individual suggests that they did not simply lock into a fixed, deterministic
action sequence.

For each participant, each queue length, and each of the $\tau$ positions in a
queue, we compute the fraction of episodes in which the participant defects at
the given position. We average these proportions across participants and then
compute empirical hazard curves. Figure~\ref{fig:results1}b shows hazard curves
for each of the six queue lengths and the two $\rho$ conditions. The
$\rho=1.50$ curves are lighter and are offset slightly to the left relative
to the $\rho=1.25$ curves to make the pair more discriminable. The Figure
presents both human data---asterisks connected by dotted lines---and simulation
results---circles connected by solid lines. Focusing on the human data for the
moment, initial-defection rates rise slightly with queue length and are greater
for $\rho=1.25$ than for $\rho=1.50$. We thus see robust evidence that 
participants are sensitive to game conditions. 

To model the population data, we set the DGMDP parameters ($\Theta_\text{task}$) based on
the game configuration. We obtain least-squares fits to the four agent
parameters ($\Theta_\text{agent}$): discount rate $\gamma=0.957$, initial and
delta \bias spreads $\sigma_1=81.3$, and $\sigma = 21.3$, and effort cost
$\re = -52.1$. The latter three parameters can be interpreted using the scale
of the SS reward, $\rss = 100$ points. 
%\mcm{CAN SAY MORE ABOUT INTERPRETING THESE PARAMETERS IF THERE IS SPACE}
Although the model appears to fit the pattern of data quite well, the model has
four parameters and the data can essentially be characterized by four
qualitative features: the mean rate of initial defection, the modulation of the
initial-defection rate based on queue length and on $\rho$, and the curvature
of the hazard function. The model parameters have no direct relationship to
these features of the curves, but the model is flexible enough to fit many
empirical curves. Consequently, we are cautious in making claims for the
model's validity based solely on the fit to Experiment 1. We note, 
however, that we investigated a variant of the model in which \bias is
uncorrelated across steps, and it produces qualitatively the {\em wrong} 
prediction: it yields curves whose hazard probability depends only on the 
steps to the LL reward. In contrast, the curves of the correlated-\bias 
account depend primarily on the distance from the initial state, $t$, but
secondarily on distance to the LL reward, $\tau-t$.
% MIKE: Can delete some of the above paragraph for space

\subsection*{Experiments 2 and 3: Modulating Effort}

To obtain additional support for the theory, we modified the queue-waiting game
such that players had to work harder and experienced more frustration in
reaching the front of the long queue. By increasing the required effort, we
may test whether model parameters fit to Experiment 1 will also fit new data,
changing only the effort parameter, $\re$. The long queue's dynamics were
modified to increase the required effort.
Instead of advancing deterministically every clock tick as in Experiment 1,
the long queue advanced in an apparently random fashion on half the ticks.
To move with the queue, the player needed to press the advance key every tick,
thus requiring exactly two keystrokes for each action in the
game FSM (Figure~\ref{fig:FSMmain}b). The game clock in Experiment 2 updated twice
as fast as in Experiment 1 (1000 msec versus 2000 msec); consequently,
the overall timing was unchanged. We tested only reward-rate ratio $\rho=1.50$.

Figure~\ref{fig:results1}c shows hazard curves for Experiment 2. Using
Experiment 1 parameter settings for $\gamma$, $\sigma_1$, and $\sigma$, we fit
only the effort parameter, obtaining $\re = -99.7$, which is fortuitously
twice the value obtained in Experiment 1. Model fits are superimposed over
the human data. To further test the theory's predictive power, we froze all
four parameters and ran an Experiment 3 identical to Experiment 2 except that
we introduced a smattering of 50 and 75 point bonuses along the path to the LL 
(see example in Figure~\ref{fig:queuewaiting}b). We also reduced the
front-of-queue reward such that the reward-rate ratio $\rho=1.50$ was attained when 
traversing the entire queue. Using the fully constrained model from
Experiment 2, the fit obtained for Experiment 3 was quite good
(Figure~\ref{fig:results1}d). The model may slightly underpredict
long-queue initial defections, but it captures the curvature of the hazard
functions due to the presence of bonuses.

\subsection*{Experiment 4: Customized Bonuses to a Subpopulation}

In Experiment 4, we tested the effect of bonuses customized to a subpopulation.
To set up this Experiment, we reviewed the Experiment 2 data to
examine inter-participant variability. We stratified the 30 participants in
Experiment 2 based on their mean reward rate per action. This measure reflects
quality of choices and does not penalize individuals who are slow. With a
median split, the {\em weak} and {\em strong} groups have average reward rates
of 103 and 132, respectively. Theoretically, rates range from 0 (always
switching between lines and never advancing) to 100 (deterministically
selecting the short queue) to 150 (deterministically selecting the long queue).
We fit the hazard curves of each group to a customized $\gamma$, leaving
unchanged the other parameters previously tuned to the population. We obtained
excellent fits to the distinctive hazard functions with $\gamma_\text{strong} =
0.999$ and $\gamma_\text{weak}=0.875$.

\begin{figure}[tb]%[100]%[bt]
  \centering
  \includegraphics[width=6in]{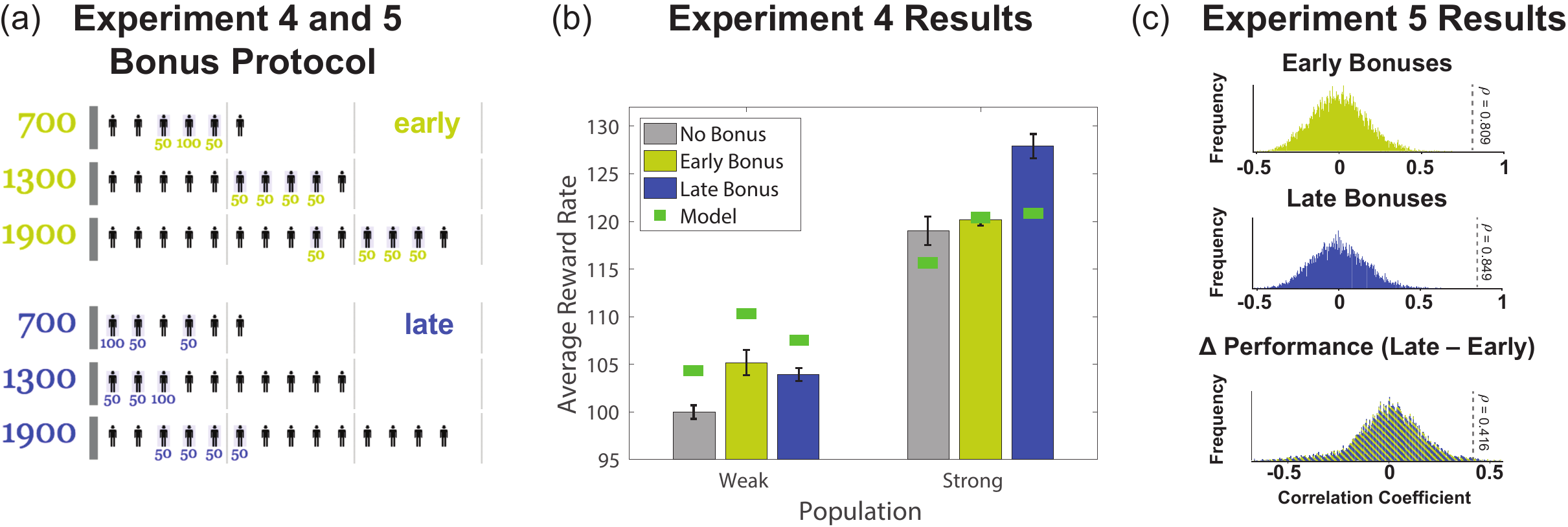}
  \caption{
  (a) Experiments 4 and 5: Model-predicted optimal bonus 
  sequences, with early (yellow) and late (blue) bonuses for weak and
  strong participants, respectively. (b) Experiment 4: Average reward rate 
  for weak and strong subpopulations and three bonus
  conditions. Error bars are $\pm1$ SEM, corrected for between-subject
  variance \protect\citep{MassonLoftus2003}.
  (c) Experiment 5, Pearson correlation of individual participant's reward rate and model prediction.
  The vertical dashed lines indicate the correlation when parameters estimated from an individual's no-bonus condition
  performance is used to  predict that individual's bonus-condition performance. The distributions reflect shuffled
  parameters, i.e.,  when model parameters fit to one individual are used to predict another individual's performance.
  }
  \label{fig:results45}
\end{figure}

We then optimized bonuses for each group for various line lengths. As in
Figure~\ref{fig:optbonus}c, we searched over a bonus space consisting of all
arrangements of up-to four bonuses, each worth fifty points, allowing multiple
bonuses at the same queue position.\footnote{We avoid the interest-accrual
setting for bonuses in this iterated task because it could lead to variable reward 
rates among episodes. Reward rate must be constant across episodes to validate
treating an iterated version of the DGMDP in Figure~\ref{fig:FSMmain}a
(see Figure~\ref{fig:FSMsupp}) as equivalent to the one-shot DGMDP in Figure \ref{fig:FSMmain}b.}
We subtracted 200 points
from the LL reward, maintaining a reward-rate ratio of $\rho=1.50$ for
completing the long queue. We constrained the search such that no mid-queue
defection strategy would lead to $\rho>1$. A brute-force optimization yields
bonuses {\em early} in the queue for the weak group, and bonuses {\em late} in
the queue for the strong group (Figure~\ref{fig:results45}a).

Experiment 4 tested participants on three line lengths---6, 10, and 14---and 
three bonus conditions---early, late, and no bonuses. (The no-bonus case was as in Experiment 2.) The 54 participants who completed
Experiment 4 were median split into a weak and a strong group based on their
reward rate on no-bonus episodes only. Consistent with the model-based optimization,
the weak group performs better on early bonuses and the strong group on late
bonuses (the yellow and blue bars in Figure~\ref{fig:results45}b). Importantly,
there is a $2\times2$ interaction between group and early versus late bonus
($F(1,51)=11.82$, $p=.001$) indicating a differential effect of bonuses on the
two groups. Figure~\ref{fig:results45}b also shows model predictions based
the parameterization determined from Experiment 2. The model has a perfect
rank correlation with the data, and correctly predicts that both bonus
conditions will facilitate performance, despite the objectively equal
reward rate in the bonus and no-bonus conditions. That bonuses will
improve performance is nontrivial: the persistence induced by the bonuses 
must overcome the tendency to defect because the LL reward that can
be obtained in the bonus condition is reduced to compensate for the bonuses.
%\mcm{I removed the following because I couldn't figure out what it meant: (as we observed in Experiment 1 with $\rho=1.25$ versus $\rho=1.50$)}

\subsection*{Experiment 5: Predicting Individual Bonus Sensitivity}

Whereas Experiment 4 customized bonuses to a subpopulation, Experiment 5 focuses on individuals.
To avoid the possible confound of intermixing of bonus and no-bonus trials, the experiment was divided into two phases:
a four minute phase with no bonuses and a 7 minute phase with the early and late bonus structures 
used in Experiment 4. 
%Line lengths were as in Experiment 4. 

We separately fit the $\Theta_\text{agent}$ parameters
%\mcm{Shruthi-is this correct?}
to the data from each participant in phase 1 and then used the parameterized model to predict average reward rate 
in the two bonus conditions of phase 2. The model obtains correlations with individuals' 
early- and late-bonus reward rates of 0.81 and 0.85, respectively. When parameters fit to one individual are used to predict 
performance of another randomly drawn individual (\emph{shuffled parameters}), the median correlation
drops to -0.004 and -0.008.  The top and middle panels of Figure~\ref{fig:results45}c show the correlation
distribution over 10,000 shuffles. The matched-parameter model is clearly an outlier: 
none of the 10,000 shuffles yields correlations higher than the one we observe for the matched-parameter model.
Thus, the parameters inferred from the no-bonus phase are able to predict a specific individual's response to the
presence of bonuses.

A critical test of the theory is whether it can anticipate which bonus structure is superior for an individual.
The matched-parameter model obtains a correlation of 0.42 with the reward-rate difference between early and late
bonuses, versus a median of 0.005 for shuffled parameters.  The model predictions
are far better than one would expect without insight into an individual's decision making processes
(Figure~\ref{fig:results45}c, bottom panel): only 30 of 10,000 shuffled parameters yield a correlation as large as the
matched-parameter model (i.e., $p=.003$). One should not be disappointed that the model cannot perfectly predict
the relative advantage of early versus late bonuses: model-parameter and reward-rate estimates are based on
only \emph{three} minutes of data collection apiece. Consequently, the resulting model predictions and
dependent measures are intrinsically noisy which bounds the maximum correlation.

\section*{Discussion}

%\color{blue}
In this article, we developed a formal theoretical framework to modeling the
dynamics of intertemporal choice. We hypothesized that the theory is suitable
to modeling human behavior. We obtained support for the theory by
demonstrating that it explains key qualitative behavioral phenomena 
and predicts quantitative outcomes from a series of behavioral experiments.
Although our first experiment merely suggests that the theory has
the flexibility to fit behavioral data post hoc, each following experiment used
parametric constraints from the earlier experiments, leading to strong
predictions from the theory that match behavioral evidence. The theory allows
us to design incentive mechanisms that steer individuals toward better outcomes,
3), and we showed that this idea works in practice for customizing
bonuses to subpopulations and individuals playing our queue-waiting game. 
%The theory and the behavioral evidence both show a non-obvious and non-intuitive statistical
%interaction between the subpopulations and various incentive schemes. 
Because the theory has just four free parameters, it is readily pinned down to
make strong, make-or-break predictions. Furthermore, it should be feasible to
fit the theory to individuals as well as to subpopulations. With such
fits comes the potential for maximally effective, truly individualized
approaches to guiding intertemporal choice.

The contributions of our work can be appreciated by contrast with the recent work of \citet{Lieder2019},
which is also aimed at using formal theories of decision making to design incentives. Their incentives
take the form of ``breadcrumbs'' (our term) that reduce the cognitive effort required to attain optimal
performance. The approach depends on being able to restructure environments by manufacturing \emph{pseudo-rewards} that need to be interpreted by participants as if they are actual rewards---often expressed in dollars---but which are not actually paid out.  This approach is quite sensible in
tasks where individuals procrastinate to avoid effort.
However, in a delayed gratification task, 
this approach is effectively like promising a participant in a retirement plan that they will receive 
a 65" flat screen TV if they deposit funds in their retirement account and then not actually delivering 
it. Pseudo-rewards may work in cognitive tasks where participants are seeking breadcrumbs to 
follow, but unfulfilled promises will quickly lose their appeal in a delayed-gratification 
task.\footnote{Throughout, we assume that bonuses carry a cost, but even in rare situations
where cost-free incentives can be identified \citep{woolley2016}, overuse
can reduce their value and consequently, selective placement of incentives 
still matters.}
%  ADRIAN COMMENT
%In our experiments and simulations, there is a tension in offering bonuses
%or incentives to encourage adherence to long-term goals because doing so
%reduces the long-term rewards. Not always going to be the case
%(http://faculty.chicagobooth.edu/ayelet.fishbach/research/WF\%20JCR\%20Harnessing.pdf).
%(3) If appropriate, I do think the conclusion could be a really nice place to bring up the idea of incentives that don’t require reducing the LL reward—I don’t know what those might be, but I’m thinking along the lines of the “For the Fun of It” paper I forwarded the other day (http://faculty.chicagobooth.edu/ayelet.fishbach/research/WF\%20JCR\%20Harnessing.pdf). Building on this:
%\newline
%“With high discounting, this particular simulation yielded modest (? 10\%) improvement in an individual’s expected accumulation. Bonuses are awarded only when the gain from encouraging persistence beats the loss of interest on a distribution.”
%\newline
%…the idea would be that any increase in persistence from a non-monetary incentive will always beat the loss of interest (because there is no loss of interest). I think it would be cool to reemphasize the tension inherent in encouraging adherence to LL goals by reducing LL rewards, then offer a potential alternative.
Another contrast between our work and that of \citet{Lieder2019} is our focus on modeling individuals by
fitting model parameters using a behavioral assessment and then optimizing incentives to the individual.
This exercise places strong demands on the model.

The next step in our research program is to demonstrate utility in 
incentivizing individuals to persevere toward long-term goals on the time scale of months and years such as 
losing weight or saving for retirement. It remains uncertain whether 
intertemporal choice on a long time scale has the same dynamics as
on the short time scale of our queue-waiting game. However, our modeling framework is
in principle scale invariant, and the finding
that reward-seeking behavior on the time scale of eye movements can be
related to reward-seeking behavior on the time scale of weeks and months 
\citep{Shadmehretal2010,WolpertLandy2012} leads us to hope for scale invariance.

%\shs{Discussion - add opportunity cost to the discussion as away to talk about sequential dependencies between episodes}
%\mcm{one possible alternative to my paragraph about Lieder above would be to shorten the paragraph and just address
%the fact we're fitting the model to individuals, followed by a paragraph in which we discuss many possible extensions and
%task settings, including: (1) the case where bonuses are free (which includes Leider et al.), (2) shruthi's opportunity costs,
%and maybe (3) more detail on some of the scenarios adrian found interesting.}
%
% MIKE OMITTED THIS COMMENT OF ADRIAN'S BECAUSE I'M NOT SURE WHAT TO SAY
% our model has no way of distinguishing 1300+200 vs. 1500
%(2) One interesting thing I noticed that might not be appropriate for a “conclusion”: the reward-rate ratio affects initial decisions (initial defection rates), and bonuses affect adherence to these decisions (at least for those in the “weak” group). Because bonuses involve reducing the amount of the LL reward (and thus reducing the reward-rate ratio), they may reduce the probability of an agent entering the LL line—and bonuses can’t keep agents in lines they never enter. Unless I missed it in expt 4, I don’t think we tested a no-bonus line with a 1500 LL reward vs. a bonus line with a 1300 LL reward and 200 bonus points.
%
\section*{Methods}

\noindent
\textbf{Experimental protocol}. The experiments were conducted with informed consent approved by from the University of Colorado's Institutional Review Board. Participants were recruited using the Amazon Mechanical Turk website and were compensated for their time at the rate of 8\$ per hour. Five experiments were conducted all of which were based on the same basic protocol. Subjects were provided instructions to play the 'Queue Waiting game' in which they controlled the position of a player (in red, Figure \ref{fig:queuewaiting}a). When in the waiting area, subjects could choose to move their player to either a short queue with no waiting time but a reward amount of $100$ points, or a longer queue with varying lengths, $\tau$, but proportionally larger rewards, $100 \tau \rho$ points. Each experiment had a fixed duration for which subjects repeatedly performed the queue waiting task. The longer queue had a higher rate if reward, $\rho$, making it the more rewarding choice over the course of the game. The queue lengths, as well as the reward rate of the longer queue proportional to the short queue, were varied in each experiment to test varying aspects of subjects' behavior.

\noindent
\textbf{Bonus presentation}. The protocol was designed with the expectation that, to varying degrees, subjects would commit to the longer queue at the beginning and defect to the shorter one before obtaining the larger reward. Therefore, as described in the main text, we designed incentives to help individuals persist with their choice of the longer line. While the theory behind the incentive design has been described earlier, here we describe its appearance in the protocol. Bonus points were systematically placed in selected positions in the queue (Figure \ref{fig:queuewaiting}b) and correspondingly reduced from the final reward, in effect shifting a small portion of the reward earlier in time. Subjects received audio-visual stimuli corresponding to the bonus rewards which changed with the bonus magnitudes as well; the auditory sounds were all variations of a cash register "Ka-Ching" sound. 

\noindent
\textbf{Data analysis and metrics}.
In our analyses of player behavior we found that at the start, players are learning the game actions and at the end, players may not have sufficient time to traverse the long queue and defection is the optimal strategy. Therefore, we remove the first and last thirty seconds of play. To measure their behavior we compute an empirical hazard rate (theoretically defined as in Equation \eqref{eq:hazard}); for each participant, each queue length, and each of the $\tau$ positions in a queue, we compute the fraction of episodes in which the participant defects at the given position. We average these proportions across participants and then compute empirical hazard rate per line position per subject. When visualizing the theoretical or empirical hazard rates with respect to line position, we are able to generate hazard curves, using which we can determine how well the model fits empirical data as in Figure \ref{fig:results1}b---d. 

Across all the experiments, participants are paid at a rate of \$8.00 per hour and are awarded a score-based bonus. If they fail to act, they are warned after 7 idle seconds and rejected after 14 seconds. They are also rejected if they de-focus their browser twice, with a warning message after the first time. Specifics regarding the protocol are described for each experiment below. 

In each experiment, the first thirty seconds of game play are discarded to allow participants to figure out the
game, and the last thirty seconds are discarded due to the possibility of greedy end-of-game strategies.
In Experiment 5, the last 30 seconds of phase 1 game play are also discarded.

\vspace{8pt}
\noindent
\textit{\textbf{Experiment 1:}} In this experiment, two values of the reward rate ratio, $\rho \in \{ 1.25, 1.50\}$, were tested to determine participants' sensitivity to the conditions of the game. Six lengths were tested across multiple iterative episodes for the long queue uniformly drawn from $\{4, 6, 8, 10, 12, 14\}$. At each update of the game state which happened every 2 seconds, both queues advanced. The player advanced along with the queues if subjects pressed the left arrow key just before the game state updated. Both queues advanced deterministically on each update. Forty-one participants were recruited, twenty for the $\rho = 1.50$ condition and twenty-one for the $\rho = 1.25$ condition. 

To compare the empirical data on participants' behavior to the model's predictions, we simulated the model by setting the task parameters $\Theta_{task}$ in our \textit{DGMDP} based on our game configuration and obtained least squares fits for the parameters that determine subjective behvaiour $\Theta_{agent}$, namely the discount rate $\gamma$, the spread in \bias distribution $\sigma$ as well as the initial value at the first time step $\sigma_1$, and effort cost $\re$.

% In Experiments 1-3, the
% long-queue length $\tau$ is uniformly drawn from $\{4, 6, 8, 10,
% 12, 14\}$ for each epsiode. 
\vspace{8pt}
\noindent
\textit{\textbf{Experiment 2:}} For this experiment, moving the player through the line was made more effortful. The game state was updated every second, twice as fast as in experiment 1. However, to move the player through the queue, while every game state update required a key stroke, only one would result in a movement requiring two keystrokes per action. Therefore, advancing one position in the queues required two keystrokes as opposed to one, with the longer queue advancing pseudorandomly ensuring that an equal number of advances got the player to the end of the queue. The same set of lengths for the long queue as in experiment 1 was experienced by participants in experiment 2. Only one reward rate ratio $\rho = 1.50$ was tested in this experiment and thirty participants were recruited to perform the task.

The effort parameter from the model $\re$ was obtained by fitting the model to the data while using the fits for the other parameters as determined from experiment 1 ($\gamma$,$\sigma$,$\sigma_1$).

% increased effort, no bonus
% 30 participants
% In Experiments 2-5,
% game state updates at 1000 msec intervales,
% short queue advances deterministically on each update,
% long queue advices in a pseudorandom fashion ensuring that the
% minimium numbers of steps to obtain reward for a long queue of length $\tau$ is exactly $2 \tau$,
% and the minimum number of steps for a short queue (length 1) is $2$.

% 3.0 = experiment1, control
%   NIPS2016: EXPERIMENT 2 (increased effort, no bonuses)
%   30 subjects
\vspace{8pt}
\noindent
\textit{\textbf{Experiment 3:}} This experiment was run to test the predictive power of a fully-constrained model from experiment 2 on unseen data. Further, we added small bonus rewards for making it to certain positions (worth 50 or 75 points) in the long queue which were correspondingly subtracted from the final reward in the queue to maintain the reward rate ratio of $\rho = 1.50$ if the entire queue was traversed. The fully constrained model's hazard rates were compared with empirical performance for the population in this experiment with the game parameters (bonuses and final reward) adjusted accordingly. Length of the longer queue was sampled from the same set of six lengths as in the previous two experiments. Thirty participants were recruited for this task. 
% Thirty participants total were tested.
% In this Experiment, bonuses were awarded as the player reached certain positions in the queue.
% HOW BIG AND HOW MANY AND WHERE FOR EACH LINE LENGTH.
% PROBABLY HAVE TALK SLIDES WITH THE BONUSES AND POSITIONS -- put in Supplementary Information
% 4.0 = experiment1, bonus
%   NIPS2016: EXPERIMENT 3 (increased effort, bonuses)
%   30 subjects

\vspace{8pt}
\noindent
\textit{\textbf{Experiment 4:}} This experiment was designed to determine the effects of customized bonuses to sub-populations. Bonus schemes were designed based on data from experiment 2 within which subjects' performance was stratified and categorized into two groups---\textit{strong} and \textit{weak} groups based on their earned average reward. Models were fit to the two groups separately and bonuses were then optimized to improve performance based on simulations for the two sub-populations. Bonuses awarded were subtracted from the final reward as in experiment 3 to maintain the reward rate. Brute force optimization yielded a strategy that provided early bonuses for the \textit{weak} group and late bonuses to the \textit{strong} group. 
In this experiment, three lengths were used for the longer queue, $\tau \in \{6, 10, 14\}$. The three queues were presented in three conditions---the no bonus, early bonus and late bonus conditions. Fifty-four participants were recruited to perform this experiment. When analyzing the data subjects were similarly grouped into strong and weak categories based on a median-split on their empirical reward rate in the no-bonus condition. Their performances in the early and late bonus conditions were then compared using a two-way repeated measures ANOVA to test the model's hypothesis for the sub-populations' performance on the two bonus conditions. 
% 53 participants. Control condition intermixed with two bonus
% conditions (early, late). Three line lengths (6, 10, 14).

% Median split led to 26 strong and 27 weak participants.

% Removed effort penalty when a bonus is awarded.
%   %- in simulation, replaced perceived_bonus_factor with suppression of effort penalty when bonus is awarded
%   %NIPS2016: EXPERIMENT 4 

% 13.0 experiment9
%   - experiment with a no-bonus and bonus phase separated
%   - Mike's summary: The experiment is divided into 2 phases. In Phase 1 (4 minutes), the subject has to choose between a short line and a line of length {6,10,14}; there are no bonuses. In Phase 2 (7 minutes), the subject has to make the same choices, but half the trials have early bonuses and half have late bonuses. The no bonus / bonus conditions are confounded with phase ordering, so we can't make a big deal out of differences between how subjects behave in the no bonus condition relative to the bonus conditions (e.g., more defections overall in the bonus conditions) because those differences may be due to the fact that subjects are fatigued. We also have a confound with practice effects, though we expect most of the practice benefits to occur in the first 30 sec of the experiment (which is discarded for data analysis).

\vspace{8pt}
\noindent
\textit{\textbf{Experiment 5:}} The final experiment was designed to determine if the model could predict performance of individuals in specified bonus conditions. The experiment had two phases, the first being a \textit{no bonus} phase followed by a bonus phase with both early and late bonuses. Lengths used for the longer queue were the same as in experiment 4. Forty subjects were tested in this experiment, but data from one had to be excluded given abnormally long date update times indicating a problem with the game run on their end. The \textit{no bonu}s phase was four minutes long while the \textit{bonus} phase was seven minutes long. The game state was updated every second. 

For each individual participant, the DGMDP was fit to their \textit{no bonus} phase to determine each individual set of parameters $\Theta_{agent}$. Based on the fit parameters per individual, their performance in the early and late bonus phase was predicted and compared to their empirical performance. To determine the efficacy of the fit, each individual participant's fitted model was used to predict performance of of another randomly selected subject to ensure that predictions were not just random. For each bonus type, early and late, predicted and actual performance was compared within subject along with 250 shuffled pair comparisons to ensure performance was above random chance. Further, to ensure that the model was not just scaling performance of good or bad subjects in the bonus conditions, we also determined how good it was in determining the difference between each individual's performance in the early versus the late bonus conditions. Once again, the predicted and actual data pairs were shuffled 250 times to compare performance to random chance.

\newpage
\acks{This research was supported by NSF grants DRL-1631428, SES-1461535,
SMA-1041755, and seed summer funding from the Institute of
Cognitive Science at the University of Colorado. We thank John Lynch for initial
discussions that led to this collaboration, Ian Smith and Brett
Israelson for assistance in the design and coding of experiments,
and Pradeep Shenoy and Doug Eck for
helpful comments on an earlier draft
of the manuscript.}

\vskip 0.2in

\bibliography{references}
\bibliographystyle{unsrtnatapa}%apalike

\newpage
\setcounter{figure}{0}

\centerline{\LARGE\bf Supplementary Information}\vspace{0.7ex}

\section*{\large{Modelling}}
\subsection*{\Bias} 
Critical to our model is the random walk in \bias, $w$ (Equation~\ref{eq:qdefect} in the
main article). As we stated in the main article, it is essential for the model predictions
that $w$ has temporal autocorrelations. Ideally, we might have selected pink ($1/f$) noise 
rather than brown (Gaussian) noise for its scale invariant property. Our model should
be capable of explaining behavior in tasks where defections happen on the time scale of seconds
(e.g., the line-waiting game) to years (e.g., retirement planning); scale invariant pink
noise facilitates such scale invariance. However, the Gaussian formulation is mathematically 
convenient and facilitates simulations and our approximations.

Depending on the time scale, the \bias might conceivably reflect fluctuations in life stress,
sleep deprivation, mood, hunger, or cognitive load. However, we resist attaching an
association between $w$ and these cognitive factors, as evidence suggests that factors such as
hunger affect delay discounting \citep[e.g.,][]{skrynka2019}. We also resist considering $w$ 
to relate directly to willpower, resolve, impulsivity, or grit: we argue that these psychological constructs 
emerge from the operation of a complex decision-making system rather than being primitive mechanisms
like the fluctuations in $w$. Also, these constructs operate on a much slower time scale than 
the second-to-second fluctuations we model in our line-waiting experiment and impulsivity and
grit are considered  enduring personality traits not  a time-varying state \citep{Duckworth2007}.

We rejected several alternative forms of noise.
\begin{enumerate}
    \item 
An obvious possibility, mentioned previously, 
is treating the discounting rate as a random variable. However, our goal is to propose a 
model that could be considered scale invariant, and it seems cognitively implausible 
that discount rates fluctuate significantly on a second-by-second basis, considering that 
they are often used as a stable biomarker of individual differences \citep{montague2012}. 
\item
We also decided against using $w$ as a multiplicative  modulation, partly because the 
additive form is more amenable to analyses of the model, and partly because doing so
would predict insensitivity to scaling of SS and LL rewards.
\item
Our modeling and prediction was over a time period of minutes. Over long time periods, it may
be necessary to consider a mean-reverting diffusion process that causes $w$ to decay back
to zero in the absence of noise perturbations. We omitted decay simply to avoid an additional
parameter of the model, i.e., essentially fixing a decay rate of 0.
\end{enumerate}

For the purpose of our model, the critical decision for $w$ is that it is a random
process that cannot be  directly manipulated by executive control processes.

\subsection*{Approximating the Value Function}

Consider the shape of $V(t, w)$.
With high \bias
($w \to \infty$), the agent almost certainly persists to
the LL reward and the function asymptotes at the discounted $\rll$. 
With low \bias ($w \to -\infty$), the agent almost certainly defects and the function
approaches $\rss - w$. Thus, both extrema of the value function are linear
with known slope and intercept. 
At step $\tau$, these two linear segments exactly define the value function.
At $t < \tau$, there is an intermediate range within which
small fluctuations in \bias can influence
the decision and the expectation
in Equation~\ref{eq:qpersist} of the main article yields a weighted mixture of the two extrema, 
which is well fit by a single linear segment---defined by its slope $\at$ and 
intercept $\bt$. With $V(t,w)$ expressed as a piecewise-linear approximation, the expectation in
Equation~\ref{eq:qpersist} of the main article becomes:
\begin{equation}
% N-3 -> t-1
% N-2 -> t
% abc_{N-2} -> abc_t \at \bt \ct
% \alpha^1_{N-2} -> z^-_t \zmt * \sigma
% \alpha^2_{N-2} -> z^+_t \zpt * \sigma
\begin{split}
\mathbb{E}_{W_t |W_{t-1} = w} V(t, w_t)= &
\Phi\left( \zmt \right) (\rss - w) +
\left( \Phi\left( \zpt \right) -
    \Phi\left( \zmt \right) \right) (\bt + \at w) \\
& ~~~~~~~+ \left( 1-\Phi\left( \zpt \right) \right) \ct +
\sigma \phi(\zmt) + \sigma \at \left( \phi(\zmt) - \phi(\zpt) \right) ,
\end{split}
\label{eq:valueinduction}
\end{equation}
where $\Phi(.)$ and $\phi(.)$ are the cdf and pdf of a standard normal
distribution, respectively, and the standardized segment boundaries are 
$\zmt= \sigma^{-1}[(\rss - \bt)/(\at+1) - w]$ and 
$\zpt~=~\sigma^{-1}[(\ct-\bt)/\at-w]$. The backup is seeded with
$z_\tau^{-} = z_\tau^{+} = \sigma^{-1} (\rss - \rll - w)$ and 
$a_{\tau}=b_{\tau}=c_{\tau}=\rll$.
After each backup step, a Levenberg-Marquardt nonlinear least
squares fit obtains $a_{t-1}$ and $b_{t-1}$; $c_{t-1}$---the
value of steadfast persistence---is obtained by propagating the discounted
reward for persistence:
$c_{t-1}~=~\re + \mu_t + \gamma \ct$. 

To ensure accuracy
of the estimate and to eliminate an accumulation of estimation errors, we have
also used a fine piecewise constant approximation in the intermediate region, 
yet the model output is almost identical.

\subsection*{Posterior Estimation of \Bias}

To represent the posterior distribution over \bias at each non-defection step
in Equation~4, we initially used 
particle filters but found a computationally more efficient and stable solution
with quantile-based samples. We approximate the $W_1$ prior and $\Delta W$
with discrete, equal probability $q$-quantiles. We reject values for which
defection occurs, and then propagate $W_{t+1}=W_t + \Delta W$ which results in 
up to $q^2$ samples, which we thin back to $q$-quantiles at each step. Using
$q=1000$ produces nearly identical results to selecting a much higher density of
samples. 

\subsection*{Conditions of Equivalence of One-Shot and Iterative Delayed-Gratification Tasks}

\begin{figure}[bt]%[100]%[bt]
  \includegraphics[width=6in]{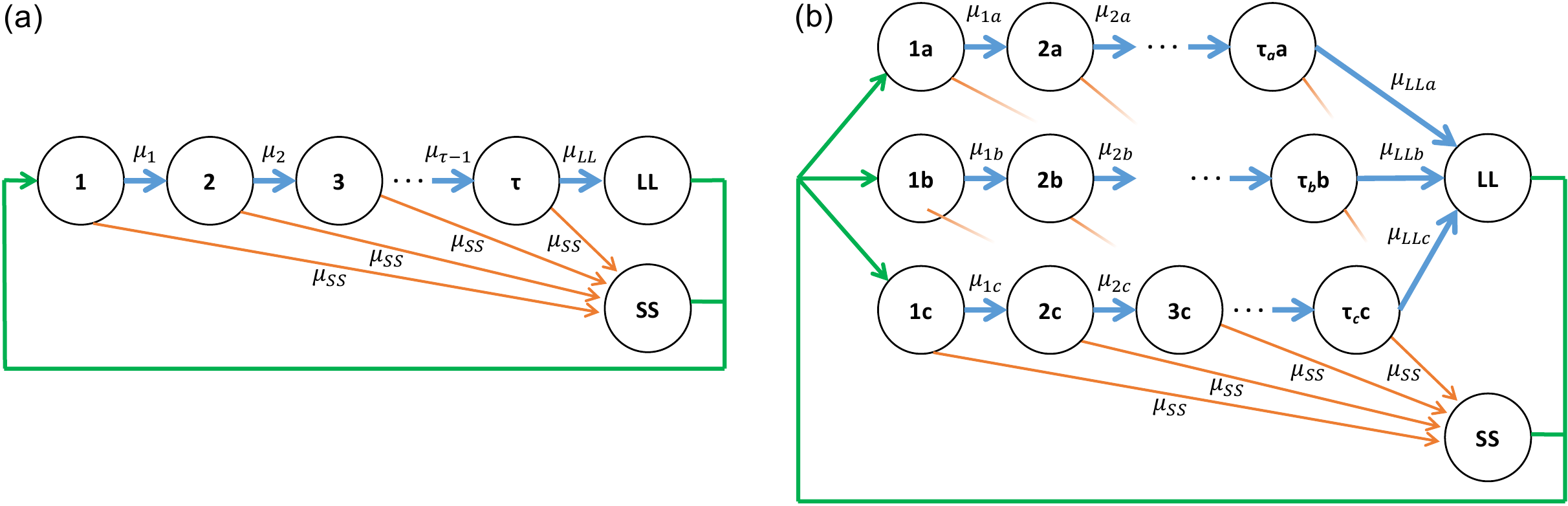}
  \caption{Finite-state environment formalizing (a) 
  the iterated  delayed-gratification task; (b) the iterated delayed-gratification task
  with variable delays and LL outcomes
  \label{fig:FSMsupp}}
\end{figure}

The one-shot DGMDP in Figure~\ref{fig:FSMmain}a of the main article can be extended to model the
iterated task, shown in Figure \ref{fig:FSMsupp}a, even when there is variability in the
reward ($\rll$) or duration ($\tau$) across \textit{episodes}, shown in
Figure \ref{fig:FSMsupp}b. Figures~\ref{fig:FSMsupp}a,b describe an
indefinite series of episodes. If the total number of episodes or steps is
constrained, as in any realistic scenario (e.g., an individual has eight hours
in the work day to perform tasks like answering email), then the state must be
augmented with a representation of remaining time. We dodge this complication
by modeling situations in which the `end game' is not approaching, e.g., only
the first half of a work day.

It is straightforward to show that the solution to the iterated DGMDP in Figure~\ref{fig:FSMsupp}b is
identical to the solution to the simpler and more tractable one-shot DGMDP in
Figure~\ref{fig:FSMmain}b of the main article under certain constraints that we describe next.
These constraints ensure that there is no interdependence among episodes, allowing a one-shot
DGMDP (Figure~\ref{fig:FSMmain}b) to serve as a proxy for an iterative task (Figure~\ref{fig:FSMsupp}b).

Consider the value function for a special case where the \bias 
does not fluctuate, i.e., $\sigma^2 = 0$ and where intermediate rewards
are not provided, i.e., $\mu_i = 0$ for $i \in \{1 ... \tau-1 \}$.
In this case, we can show that the solution to the DGMDP in Figure \ref{fig:FSMsupp}b
is identical to the solution to the DGMDP in Figure~\ref{fig:FSMmain}b of the main article.

We need to extend this result to the following more general cases, roughly in 
order of challenge:
\begin{itemize}
\item Allow for nonzero intermediate rewards
\item Allow for the case of Figure \ref{fig:FSMsupp}b where 
$\mu_\textrm{\tiny LLa}/\tau_a = \mu_\textrm{\tiny LLb}/\tau_b$
for all $a$ and $b$,
\item Allow for the case where $\sigma^2 > 0$
\end{itemize}

\subsubsection*{Proof of $\sigma^2=0$ and $\mu_i=0$ case }

In Figure~\ref{fig:FSMsupp}a, the value of state 1 is defined by the Bellman equation as:
\begin{equation}
V(1) = \max(\rss + \gamma V(1), \gamma^{\tau-1} [\rll + \gamma V(1)] )
\end{equation}
We can solve for $V(1)$ if the first term is larger:
\begin{equation}
V_\mathrm{SS}(1) = \frac{1}{1-\gamma} \rss .
\label{eq1a}
\end{equation}
We can solve for $V(1)$ if the second term is larger:
\begin{equation}
V_\mathrm{LL}(1)=\frac{\gamma^{\tau-1}}{1-\gamma^\tau} \rll .
\label{eq1b}
\end{equation}

Now consider Figure~\ref{fig:FSMmain}b of the main article, whose Bellman equation can be simplified to:
\begin{align}
V(1) &= \max \left(\sum_{i=0}^{\tau-1} \gamma^i \rss, \gamma^{\tau-1} \rll \right)\\
& = \max \left(\frac{1-\gamma^\tau}{1-\gamma} \rss, \gamma^{\tau-1} \rll \right)\\
& = (1-\gamma^\tau) \max \left(\frac{1}{1-\gamma} \rss, \frac{\gamma^{\tau-1}}{1-\gamma^\tau} \rll \right) .
\label{eq2}
\end{align}
Note that the two terms inside the max function of Equation~\ref{eq2} are
identical to the values in Equations~\ref{eq1a} and \ref{eq1b}, and thus
the value functions for Figure~\ref{fig:FSMmain}b of the main article and
Figure~\ref{fig:FSMsupp}b are identical up to a scaling  constant.

%\bibliography{references}
\subsection*{Hyperbolic discounting}
One challenge to modeling behavior with MDPs is that it is mathematically
convenient to assume exponential discounting, whereas studies of human
intertemporal choice support hyperbolic discounting
\citep{Fredericketal2002}. \citet{KurthNelson2010} have proposed a solution
to this issue by exploiting the fact that a hyperbolic function can be
well approximated by a mixture of exponentials. In our models, we found that exponential
discounting was adequate to explain behavior, but
our approach could readily be extended
in the same manner as \citet{KurthNelson2010}.
With a mixture of exponential discounting rates \cite{tiganj2017,fedus2019}, it becomes feasible to
model individual moment-to-moment variability as fluctuations in the discount
rate, which correspond to different weightings of the exponential decays.
%Cite papers by Liam, Kenji Doya, Per and Marc.}

\section*{Simulation details}
In simulations, we assume the effort cost $\re=0$ on steps when the player is served the
front-of-line reward. Similarly, we assume that $\re=0$ on any step leading to a bonus.

\subsection*{Experiment 5}

\begin{figure}[tb]%[100]%[bt]
  \centering
  \includegraphics[width=6in]{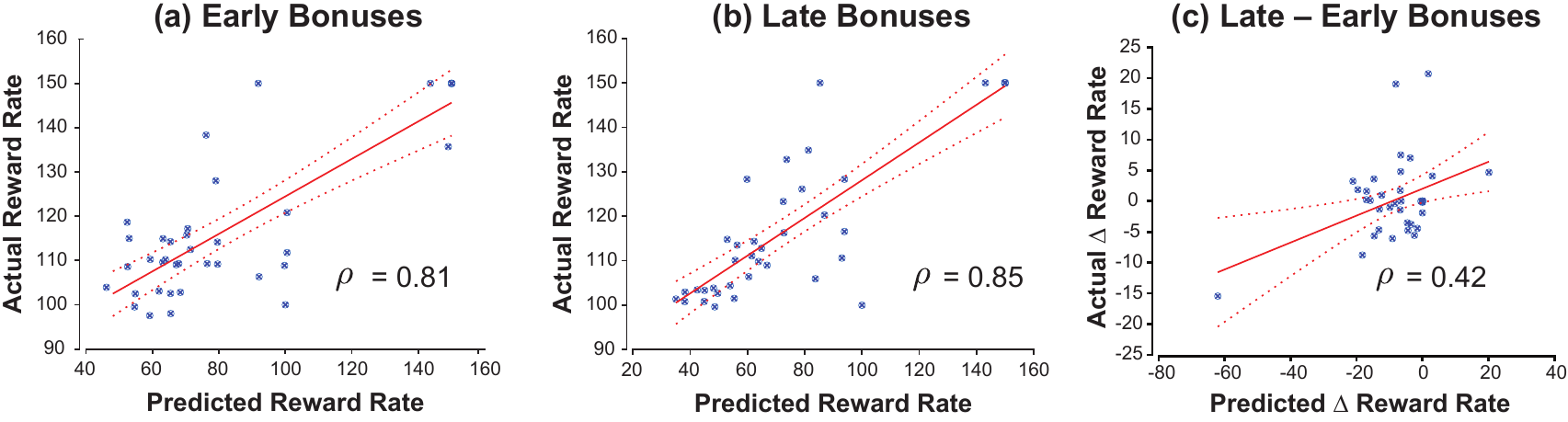}
  \caption{Experiment 5, comparing predicted and observed reward rates.
  The predicted reward rate for each individual is estimated via model simulations with model parameters fit to the individual's no-bonus phase data.
  (a) Early-bonus condition scatterplot, with regression line and confidence interval.
  (b) Late-bonus condition scatterplot.
  (c) Difference between reward rates for early versus late bonus conditions.
  }
  \label{fig:expt5results}
\end{figure}

In Experiment 5, we obtained parameters $\Theta_\text{agent}$ for each individual based on their behavior in phase 1.
We then predicted individuals' reward rate on the early and late bonus trials in phase 2. The scatterplot of
predicted and actual reward rates is shown in Figure~\ref{fig:expt5results}a,b for early and late bonuses. 
Figure~\ref{fig:expt5results}c shows the correlation of the \emph{difference} between late and early rates.
To evaluate the degree to which these parameters characterize an individual's behavior, we shuffled the assignment of
parameters to individuals. 

\section*{Experiment Instruction Phase}

\begin{figure}[H]
    % \centering
     \begin{subfigure}[b]{0.45\textwidth}
         %\centering
         \includegraphics[width=\textwidth]{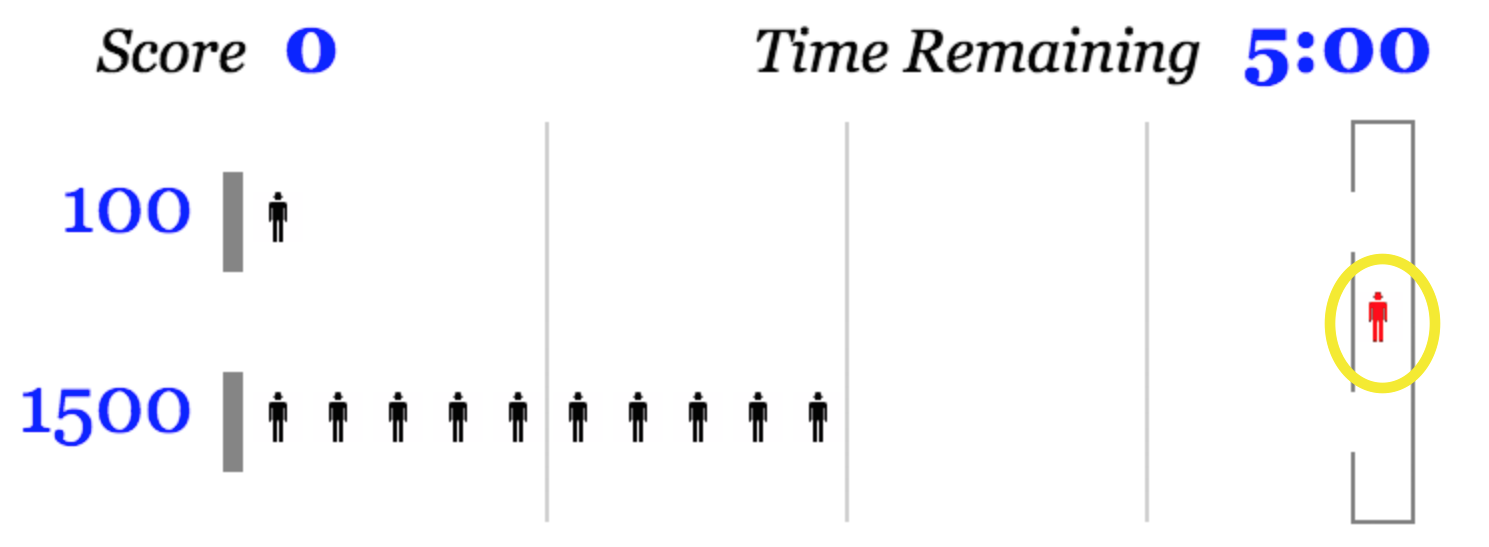}
         \caption{ }
         \label{fig:inst_screen1}
     \end{subfigure}
     \hfill
     \begin{subfigure}[b]{0.45\textwidth}
        % \centering
         \includegraphics[width=\textwidth]{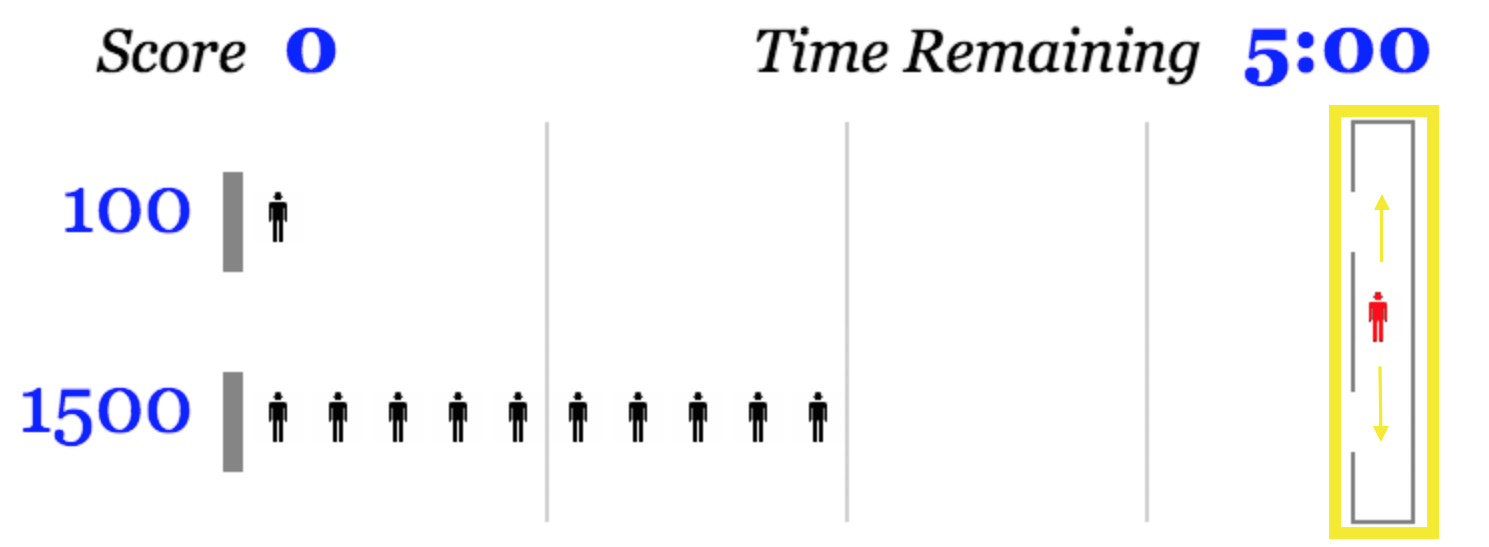}
         \caption{ }
         \label{fig:inst_screen2}
     \end{subfigure}
     \hfill
     \begin{subfigure}[b]{0.45\textwidth}
        % \centering
         \includegraphics[width=\textwidth]{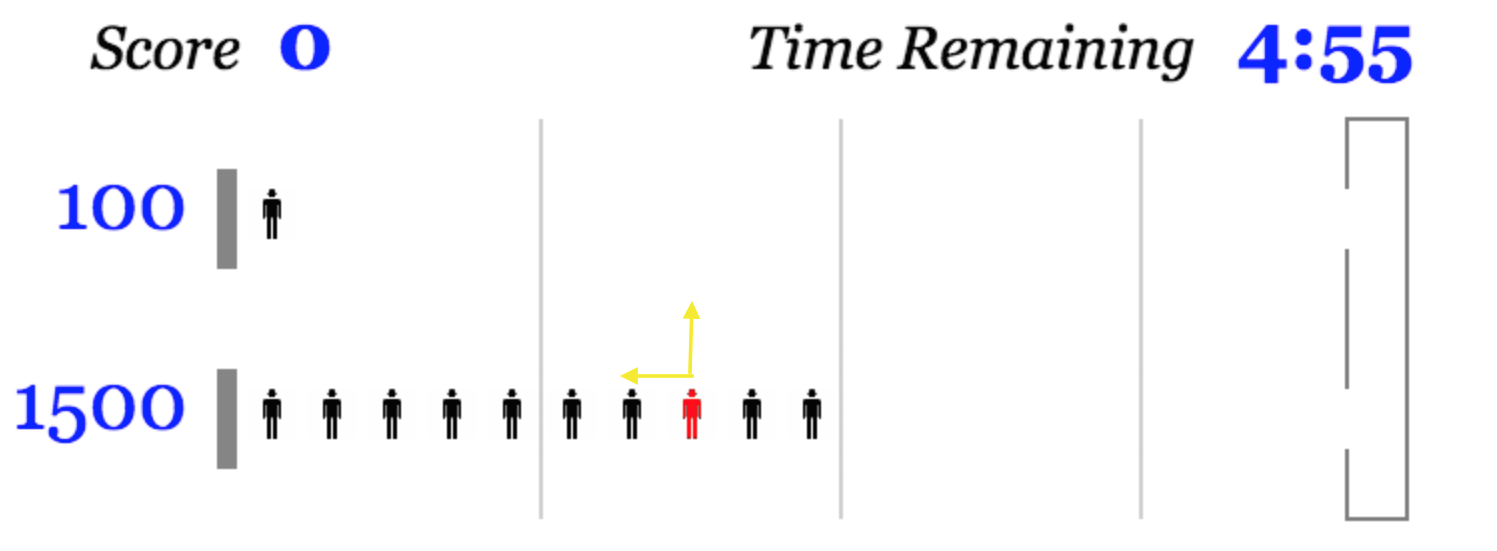}
        \caption{ }
         \label{fig:inst_screen3}
     \end{subfigure}
     \hfill
     \begin{subfigure}[b]{0.45\textwidth}
        % \centering
         \includegraphics[width=\textwidth]{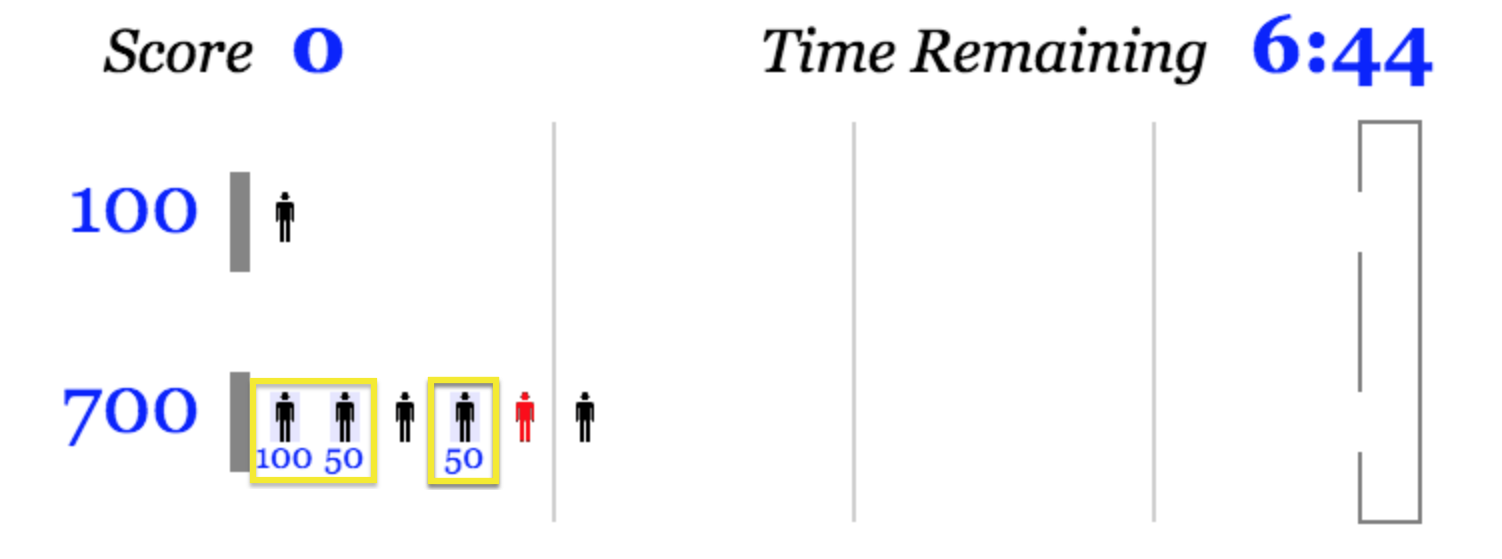}
        \caption{ }
         \label{fig:inst_screen4}
     \end{subfigure}
        \caption{Subjects were shown examples of the game screen with annotated instructions: (a) They were instructed to move their player, indicated in red to the front of the queue. (b) The up or down arrow key could be used to select the line they wanted to move trhough.(c) Subjects can advance through the queue using the left arrow key and free to switch between lines using the up or down arrow key. Once they reached the end of the queue they earned the points indicated to the left of the queues. (d) Subjects also were able to earn bonuses at certain spots in the queue as indicated under the spot.}
        \label{fig:experment_instructions}
\end{figure}

Before each experiment began, subjects were given instructions and shown corresponding examples of the game screen. In each experiment, they were  instructed that their goal was to reach the front of the queue to score the points indicated to the left of each queue, and that they controlled the stick figure in red (Fig \ref{fig:inst_screen1}). Following this they were given instructions to choose the top or bottom queue by either pressing the up or down arrow key when the player was in the waiting area (Fig \ref{fig:inst_screen2}). Once out of the waiting area, subjects were told that they could advance their player by hitting the ‘Left’ arrow key; they could defect between lines using the up or down arrow key. Finally, once they reached the front of the queue, they earned the corresponding reward which set the player back to the waiting area (Fig \ref{fig:inst_screen3}). In experiments 1--4, they were instructed that this task would be repeated for 5 minutes. In experiments 3 and 4, subjects also had the opportunity to earn bonuses for crossing certain spots in the larger cue (Fig \ref{fig:inst_screen4}). Once they passed that spot, they received audiovisual confirmation for earning the bonus reward. 

In experiment 5, subjects were informed that, after the control phase of 4 minutes, they will receive further instructions to perform the remainder of the task with the total task lasting 11 minutes. At the break between the tasks, they were informed that for the next 7 minutes they would also earn bonuses at certain spots (Fig (Fig \ref{fig:inst_screen4})) in the long line. 
In all experiments, they were informed that they had to keep playing to keep the experiment from being aborted.

\section*{Comment on related research}

\citet{Lieder2019} use the MDP framework to
propose incentive structures in which the one-step greedy action is also the long-term optimal action. In their Experiments 1 and 2, participants are given explicit instructions to follow the ``breadcrumbs'' (our term) provided by the incentive structure to attain optimal performance, thereby eliminating the cognitive effort required to choose the optimal action. In their Experiments 3 and 4, the approach gives direct guidance about what action participants should take next to make the most progress to a large payoff. In all their experiments, the more explicit the breadcrumbs are, the more effectively they are followed. For instance, in Experiment 4, the optimal incentives have positive and negative dollar values for choices an individual should prioritize and deprioritize, respectively (colored red and green, no less). In contrast, an alternative \emph{heuristic} incentive condition provides all positive dollar values (all colored green), which demands inspecting and sorting the relative magnitudes of the dollar values. The approach depends on being able to restructure environments by manufacturing \emph{pseudo-rewards} that need to be interpreted by participants as if they are actual rewards---often expressed in dollars---but which are not actually paid out. 

To contrast our work with that of \citet{Lieder2019}:
\begin{itemize}
\item
We study canonical delayed gratification task that requires patience, whereas the laboratory
tasks of Lieder et al.\ encourage procrastination or short cuts to avoid effort.
% in case reviewers think Lieder et al is related to us:
% Akira's dfn of procrastination: voluntary and irrational delay on (academic) tasks despite expecting negative consequences for the delays
% that's quite unlike our work
\item
We have shown that our model fits human behavioral data directly, rather than the indirect evidence
that Lieder et al.\ obtain by using their model to determine pseudo-rewards. To fit behavioral data, it is
necessary to make assumptions beyond those in the standard MDP framework in order to explain variability in
individual behavior.
\item
We customize our model to individuals (or subpopulations) through the observation of baseline behavior
and maximum-likelihood model-parameter fits. In contrast, \citet{Lieder2019} assume fixed parameters (e.g., discount factor, goal-abandonment probability) for all individuals.
\item
The bonuses we provide to participants are expressed in the same currency as outcomes \emph{and}
are actually awarded, and our bonus scheme is constrained such that the reward rate obtained by harvesting
bonuses cannot exceed that obtained with no bonuses. 
Further, our bonus computation ensures consistency in currency between the bonuses and the long term rewards. This structure can therefore be easily adopted to real-world tasks with food and monetary rewards. 
In contrast, fabricated pseudo-rewards are unbounded and have questionable subjective value in many 
scenarios. For example, if everyone is awarded 1000 stars, what value does a star have?
\item
Our emphasis has been optimizing incentives for an individual, and we have done the strong contrast to show
that incentives optimized for one individual are more effective than those optimized for another.
\end{itemize}

% \bibliography{references}
% \bibliographystyle{unsrtnatapa}%apalike
\end{document}